%% file: context-bench.tex
\documentclass{article}

\usepackage[preprint]{neurips_2026}

\usepackage{microtype}
\usepackage{graphicx}
\usepackage{float}
\usepackage{booktabs}
\usepackage{multirow}
\usepackage{placeins}
\usepackage{xspace}

\usepackage{amsmath}
\usepackage{amssymb}
\usepackage{mathtools}
\usepackage{amsthm}
\usepackage{siunitx}
\usepackage{listings}
\usepackage{xcolor}
\usepackage{subcaption}

\usepackage{babel}
\usepackage{hyperref}
\usepackage{caption}

\definecolor{lightcyan}{RGB}{10,110,150}
\definecolor{lighturlred}{RGB}{251,49,153}
\hypersetup{
  colorlinks=true,
  linkcolor=lightcyan,
  citecolor=lightcyan,
  filecolor=lightcyan,
  urlcolor=lighturlred
}
\usepackage[capitalize,noabbrev]{cleveref}

\usepackage{tcolorbox}
\tcbuselibrary{breakable,skins,listings}

\tcbset{
  paperbox/.style={
    enhanced,
    breakable,
    colback=black!2,        
    colframe=black!18,       
    boxrule=0.4pt,           
    arc=2pt,                 
    left=6pt,right=6pt,top=6pt,bottom=6pt,
    before skip=8pt, after skip=10pt,
    title={#1},
    fonttitle=\bfseries\small,
    coltitle=black,
    attach boxed title to top left={xshift=6pt,yshift=-2mm},
    boxed title style={
      colback=white,
      colframe=black!18,
      boxrule=0.4pt,
      arc=2pt,
      left=4pt,right=4pt,top=2pt,bottom=2pt
    }
  },
  accent/.style={
    borderline west={2pt}{0pt}{black!45}
  },
  accentblue/.style={
    borderline west={2pt}{0pt}{blue!55!black}
  },
  accentred/.style={
    borderline west={2pt}{0pt}{red!55!black}
  },
  accentgreen/.style={
    borderline west={2pt}{0pt}{green!45!black}
  }
}

\theoremstyle{plain}

\theoremstyle{definition}

\theoremstyle{remark}

\newcount\DraftStatus  % 0 suppresses notes to selves in text
\newcommand{\nbc}[3]{\ifnum\DraftStatus=1
	{\colorbox{#3}{\bfseries\sffamily\scriptsize\textcolor{white}{#1}}}
	{\textcolor{#3}{\sf\small$\blacktriangleright$\emph{#2}$\blacktriangleleft$}}
\fi}

\newcommand{\draftnote}[2]{\ifnum\DraftStatus=1
	\marginpar{
		\tiny\raggedright
		\hbadness=10000
		\def\baselinestretch{0.8}
		\textcolor{#1}{\textsf{\hspace{0pt}#2}}}
\fi}

\definecolor{todocolor}{rgb}{0.9,0.1,0.1} % Red

\newcommand{\parabf}[1]{\noindent\textbf{#1}\xspace}

\usepackage{multirow}
\usepackage{enumitem}
\usepackage{makecell}
\usepackage{colortbl}
\usepackage{pifont}

\definecolor{lightblue}{RGB}{243,248,252}
\newenvironment{summary}{
\begin{tcolorbox}[width=\linewidth, colback=lightblue, top=1pt, bottom=1pt, left=2pt, right=2pt]

}
{
\end{tcolorbox}
}

\definecolor{tablelightred}{RGB}{255,225,220}
\definecolor{tablelightblue}{RGB}{220,240,250}
\definecolor{tablelightpurple}{RGB}{230,220,245}

\definecolor{macaronpink}{RGB}{255,218,224}      % 浅粉色背景
\definecolor{macaronblue}{RGB}{173,216,230}      % 浅蓝色边框
\definecolor{macaronmint}{RGB}{189,252,201}      % 薄荷绿（用于example）
\definecolor{macaronlavender}{RGB}{230,230,250}  % 淡紫色（用于warning）

\newcommand{\benchmark}{{\textsc{ContextBench}}\xspace}

\begin{document}

% =========================
% TITLE + AUTHORS (arxiv format)
% =========================
\title{\benchmark: A Benchmark for Context Retrieval in Coding Agents}

\author{
  Han Li$^{1}$\quad
  Letian Zhu$^{1}$\thanks{These authors contributed equally as co-second authors.}\quad
  Bohan Zhang$^{1}$\footnotemark[1]\quad
  Rili Feng$^{1}$\footnotemark[1]\quad
  Jiaming Wang$^{1}$\quad
  Yue Pan$^{2}$\quad \\
  \textbf{Earl T. Barr}$^{2}$\quad
  \textbf{Federica Sarro}$^{2}$\quad
  \textbf{Zhaoyang Chu}$^{2}$\thanks{Corresponding authors. Email: \texttt{\{he.ye, zhaoyang.chu.25\}@ucl.ac.uk}}\quad
  \textbf{He Ye}$^{2}$\footnotemark[2]\quad \\
  $^{1}$Nanjing University \quad
  $^{2}$University College London \\
  \url{https://contextbench.github.io/}
}

% \author{
%   Han Li$^{1}$,
%   Letian Zhu$^{1,\dagger}$,
%   Bohan Zhang$^{1,\dagger}$,
%   Rili Feng$^{1,\dagger}$,
%   Jiaming Wang$^{1}$,
%   Yue Pan$^{2}$, \\
%   Earl T. Barr$^{2}$,
%   Sarro Federica$^{2}$,
%   Zhaoyang Chu$^{2,\ddagger}$,
%   He Ye$^{2,\ast}$ \\
%   \\
%   $^{1}$Nanjing University \quad
%   $^{2}$University College London \\
%   \\
%   \texttt{han.li.cs@smail.nju.edu.cn, \{he.ye, zhaoyang.chu.25\}@ucl.ac.uk}
% }

\date{}

\maketitle
% \vspace{-0.5cm}

% \renewcommand{\thefootnote}{\fnsymbol{footnote}}
% \footnotetext[0]{$^{\dagger}$Equal contribution as co-second authors.}
% \footnotetext[0]{$^{\ast}$Corresponding author. $^{\ddagger}$Co-corresponding author.}
% \renewcommand{\thefootnote}{\arabic{footnote}}

% =========================
% ABSTRACT
% =========================
\begin{abstract}
LLM-based coding agents have shown strong performance on automated issue resolution benchmarks, yet existing evaluations largely focus on final task success, providing limited insight into how agents retrieve and use code context during problem solving. 
We introduce \benchmark, a process-oriented evaluation of context retrieval in coding agents. \benchmark consists of \textit{1,136} issue-resolution tasks from 66 repositories across eight programming languages, each augmented with human-annotated gold contexts. We further implement an automated evaluation framework that tracks agent trajectories and measures context recall, precision, and efficiency throughout issue resolution. Using \benchmark, we evaluate four frontier LLMs and five coding agents. Our results show that sophisticated agent scaffolding yields only marginal gains in context retrieval (\textbf{``The Bitter Lesson''} of coding agents), LLMs consistently favor recall over precision, and substantial gaps exist between explored and utilized context. 
\benchmark augments existing end-to-end benchmarks with intermediate gold-context metrics that unbox the issue-resolution process. 
These contexts offer valuable intermediate signals for guiding LLM reasoning in software tasks.
% Data and code are available at: \url{https://cioutn.github.io/context-bench/}.

\end{abstract}

\section{Introduction}

% Background, benchmarks 
% Large language model based coding agents have recently demonstrated promising performance on real-world software engineering tasks such as bug fixing and feature implementation. 
Large Language Model (LLM) coding agents, exemplified by SWE-agent~\cite{yang2024sweagent},  OpenHands~\cite{wang2025openhands}, and Agentless~\cite{xia2024agentless} have recently demonstrated strong proficiency in real-world software engineering tasks, including bug fixing~\cite{jimenez2024swebench, pan2025swegym}, feature implementation~\cite{rashid2025swepolybench, deng2025nocode_bench}, and interaction with terminal-based execution environments~\cite{merrill2025terminal_bench}.
These agents are autonomous systems capable of multi-step reasoning and tool use, enabling them to solve complex, repository-scale tasks in real-world development scenarios.

% \parabf{The Gap in SWE Benchmarking.}
\parabf{Limitations of Existing Benchmarks.}
Advancements in this area are typically assessed on benchmarks such as SWE-bench~\cite{jimenez2024swebench} and SWE-bench Pro~\cite{deng2025swebenchpro}, which curate tasks from real-world GitHub issues and require agents to submit patches validated by test suites. 
However, these benchmarks prioritize end-to-end task success rates (\textit{e.g.,} Pass@k) and ignore \textit{how coding agents arrive at their solutions}. 
In particular, they overlook the evaluation of context retrieved from large codebases, which underpins the agent's reasoning.
% In particular, they fail to assess the quality of context retrieved from large-scale repositories, which forms the basis for the agent's reasoning.
Without verifying intermediate contexts, high success rates may result from trial-and-error or overfitting to specific test cases, yielding coding agents that are unreliable in practical scenarios.
We therefore argue for a context-based evaluation system that opens the black box of software tasks and addresses the key question: \textit{``How do LLM agents retrieve and use critical code context when resolving software engineering tasks?''}
\parabf{\benchmark: Beyond End-to-End Benchmarking.}
To address this question, we develop a semi-automated pipeline for curating \textit{Gold Contexts}, defined as standard sets of repository artifacts that expert developers identify as necessary to resolve a given issue.
We begin by sampling high-quality tasks from four widely used benchmarks, \textit{i.e.}, SWE-bench Verified~\cite{jimenez2024swebench}, Multi-SWE-bench~\cite{zan2025multiswebench}, SWE-PolyBench PB500~\cite{rashid2025swepolybench}, and SWE-bench Pro~\cite{deng2025swebenchpro}.
% , which are curated from real-world GitHub repositories and issues.
After deduplication via rule-based matching and embedding-based similarity detection, we select challenging tasks based on three difficulty metrics, \textit{i.e.}, agent solvability; edit scope; and edit dispersion based on gold patches.
% We then perform deduplication through both rule-based matching and embedding-based semantic similarity detection to eliminate exact and near-duplicate task instances.
% Next, we select challenging instances based on three difficulty metrics, \textit{i.e.}, agent solvability, the scope and dispersion of edits in the ground-truth patch, prioritizing tasks that exhibit high complexity or remain unsolved by existing agents.
For each selected task, human experts iteratively annotate and refine the required context in a human-in-the-loop process, ensuring compact and sufficient gold contexts for issue resolution.

% For each selected task, we adopt a human-in-the-loop procedure for gold context extraction, where expert developers annotate an initial set of candidate contexts required for issue resolution.

% To verify the sufficiency of each annotated context, we provide it to an LLM (\textit{e.g.}, GPT-5) and evaluate whether the model can generate a patch that passes the official test suite.
% If the patch passes, annotators refine the context to ensure compactness; otherwise, annotation is repeated up to two additional rounds, and only tasks achieving high inter-annotation consistency are retained.

% To verify the sufficiency of the constructed contexts, we provide them directly to an LLM (\textit{e.g.}, GPT-5) and evaluate whether the model can generate a patch that passes the official test suite under a multi-attempt evaluation setting.
% If the patch passes, annotators refine the context to ensure compactness; otherwise, annotation is repeated up to two additional rounds, and only tasks achieving high inter-annotation consistency (measured via context similarity) are retained in the final dataset.

Building on this pipeline, we introduce \benchmark, an extensive benchmark designed to evaluate the context retrieval ability of coding agents in software engineering tasks, enabling fine-grained analysis of agent behavior beyond conventional end-to-end issue resolution metrics.
\benchmark comprises  \textit{1,136} issue resolution tasks from \textit{66} code repositories across  \textit{8} programming languages.
These tasks are annotated with \textit{522,115} lines of human-verified gold contexts covering \textit{23,116} classes and functions within \textit{4,548} files.
To facilitate efficient evaluation, we also identify a Lite subset of \textit{500} tasks 
based on task difficulties. 

\begin{figure*}[t]
  \centering
  \begin{subfigure}{0.535\linewidth}
    \centering
    \includegraphics[width=\linewidth]{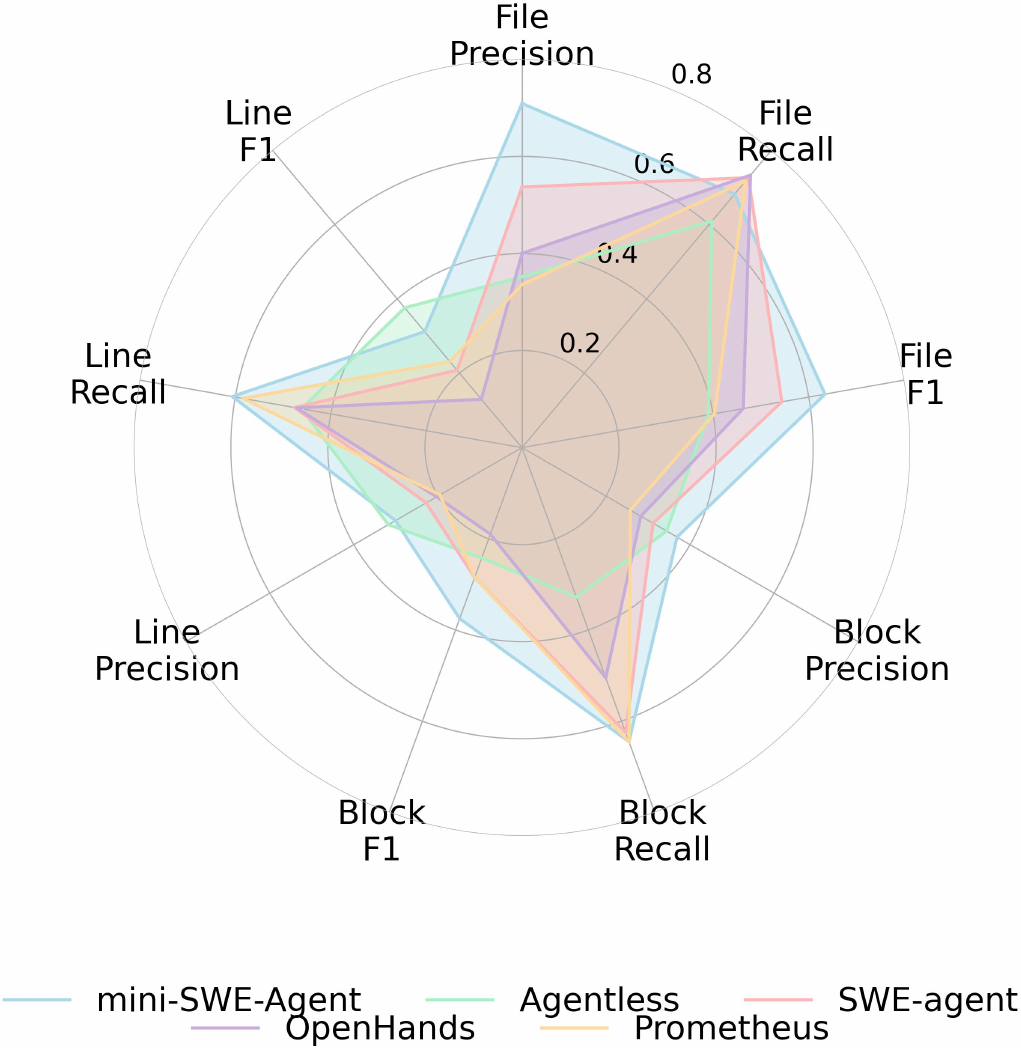}
    \caption{Coding Agent Comparison}
    \label{fig_radar_chart_sub1}
  \end{subfigure}
  \hfill
  \begin{subfigure}{0.455\linewidth}
    \centering
    \includegraphics[width=\linewidth]{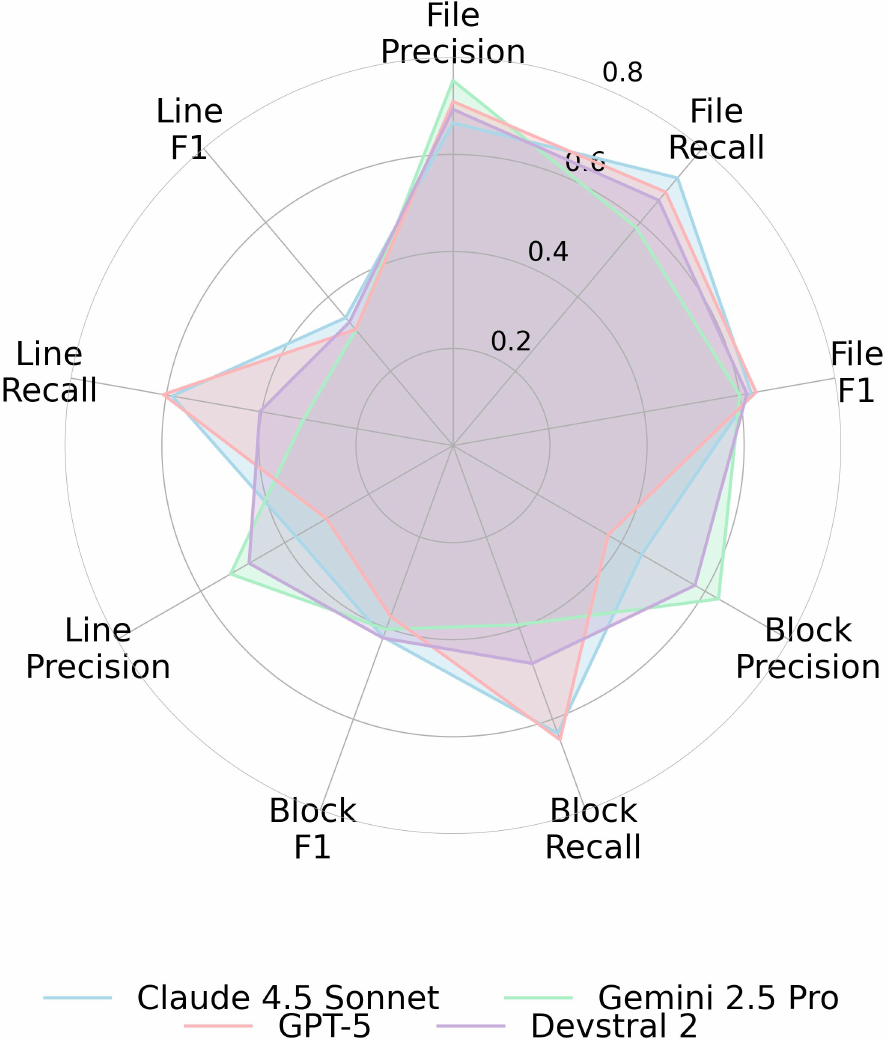}
    \caption{LLM Comparison}
    \label{fig_radar_chart_sub2}
  \end{subfigure}
  \caption{\textbf{Radar plots comparing context retrieval performance on \benchmark}. (a) Coding agent comparison and (b) LLM comparison, both evaluated at file-, block-, and line-level granularity in terms of precision, recall, and F1. Across both LLMs and agents, higher recall is consistently favored over precision.}
  \label{fig_radar_chart}
\end{figure*}

% dataset scale
% dataset statistics
% We further propose a systematic suite of metrics, implemented within an automated framework, to evaluate how effectively LLMs and coding agents retrieve code context during problem solving.
We further introduce an automated evaluation framework with a systematic suite of metrics to assess how effectively LLM agents retrieve code contexts during problem solving.
These metrics evaluate code context recall and precision at different stages of task resolving, complementing the final success rate.
Specifically, we instrument the agent’s trajectory to record all code regions it inspects. % online/offline
We then utilize the \texttt{tree-sitter}\footnote{\url{https://tree-sitter.github.io/tree-sitter/}} tool to parse the corresponding repository and map both gold and agent contexts onto a shared coordinate system that captures their file paths, AST blocks, and line ranges.
Based on this alignment, we compute recall, precision, and F1 using interval overlap at three granularities: file, block, and line levels. \autoref{fig_radar_chart} summarizes the results for four LLMs and five coding agents on \benchmark, highlighting systematic differences in retrieval behavior across models and agents.

% we perform span matching via interval overlap and report agreement at three granularities: block level, line level, and byte level. For each granularity, we compute coverage of gold regions, precision of retrieved regions, F1

% Building upon this pipeline, we introduce \benchmark, an extensive benchmark that isolates and evaluates context retrieval quality for coding agents, independently of model reasoning or execution ability. 
% \benchmark consists of hand-verified gold contexts constructed for a diverse set of real-world GitHub issues spanning bug fixes and feature requests. 
% By decoupling context retrieval from final patch correctness, \benchmark enables fine-grained analysis of agent behavior that is not possible under outcome-only evaluation.
% We further develop an automatic evaluation framework that measures context retrieval quality using coverage (recall) and precision (efficiency) metrics at multiple levels of granularity, including file, function, and code-span levels. 

%findings
% multiple agents
% multiple SOTA models
% experimental results
\parabf{Take-Aways.}
We benchmark four state-of-the-art LLMs, including GPT-5~\cite{openai2025gpt5}, Claude Sonnet 4.5~\cite{anthropic2025claudesonnet45}, Gemini 2.5 Pro~\cite{google2025gemini25}, and Devstral 2~\cite{mistral2025devstral2}, as well as five widely used coding agents, including mini-SWE-agent, SWE-agent~\cite{yang2024sweagent}, OpenHands~\cite{wang2025openhands}, Agentless~\cite{xia2024agentless}, and Prometheus~\cite{chen2025prometheus}.
Our analysis reveals several key findings: 
\ding{182}~\textbf{Sophisticated scaffolding does not necessarily lead to better context retrieval performance.} Across five coding agents, more complex retrieval scaffolds do not consistently outperform a simple baseline, as shown in~\autoref{fig_radar_chart_sub1}, suggesting potential over-engineering and echoing \textbf{``The Bitter Lesson''} of AI research.
\ding{183}~\textbf{\benchmark is challenging for state-of-the-art LLMs.} State-of-the-art LLMs struggle to retrieve effective code contexts, often covering relevant information while introducing substantial noise that undermines retrieval precision, as shown in~\autoref{fig_radar_chart_sub2}.
\ding{184}~\textbf{Recall is favored over precision in LLM context retrieval.}
All evaluated LLMs retrieve broad context to maximize coverage, introducing substantial noise and yielding limited precision and F1 gains.
\ding{185}~\textbf{Balanced retrieval improves performance and cost efficiency.}
Models that balance retrieval frequency and context granularity achieve higher Pass@1 at lower cost, while aggressive retrieval mainly increases token consumption.
\ding{186}~\textbf{Significant gaps exist between retrieved and utilized context.}
Agents often inspect gold-relevant code but fail to retain or use it in final patch generation, highlighting consolidation as a key bottleneck.

\parabf{Contributions.}
To summarize, our work provides three key contributions as follows:
% The contributions of this paper are as follows:
\begin{itemize}
    \item We construct \benchmark, containing 1,136 full instances for evaluating context retrieval in coding agents, each of which provides expert-annotated, verified gold context for assessing intermediate states of issue resolution.

    % \item We develop an automated evaluation framework that traces and aligns agent-accessed and gold contexts, enabling fine-grained measurement of retrieval coverage and precision.
% ping zhaoyang, revised above
\item \benchmark moves beyond final success-rate evaluation by enabling dynamic, process-oriented assessment across issue resolution stages through tracking and comparing agent-retrieved and gold contexts, providing an important intermediate signal for rewarding agent behavior.

\item We evaluate four frontier LLMs and five coding agents on context retrieval recall and precision, yielding new insights into how LLMs and coding agents retrieve code context during problem solving.
    
    % \item We conduct extensive experiments on state-of-the-art LLM-based agents and reveal that current systems still struggle to retrieve relevant context effectively, highlighting a key bottleneck for agentic reasoning in software engineering.

    % \item We discuss the implications of our findings for the design and evaluation of future coding agents.
\end{itemize}

\section{\benchmark: A Benchmark to Assess Context Retrieval Ability for Coding Agents}
\begin{figure*}[!t]
	\centering
	\includegraphics[width=0.98\linewidth]{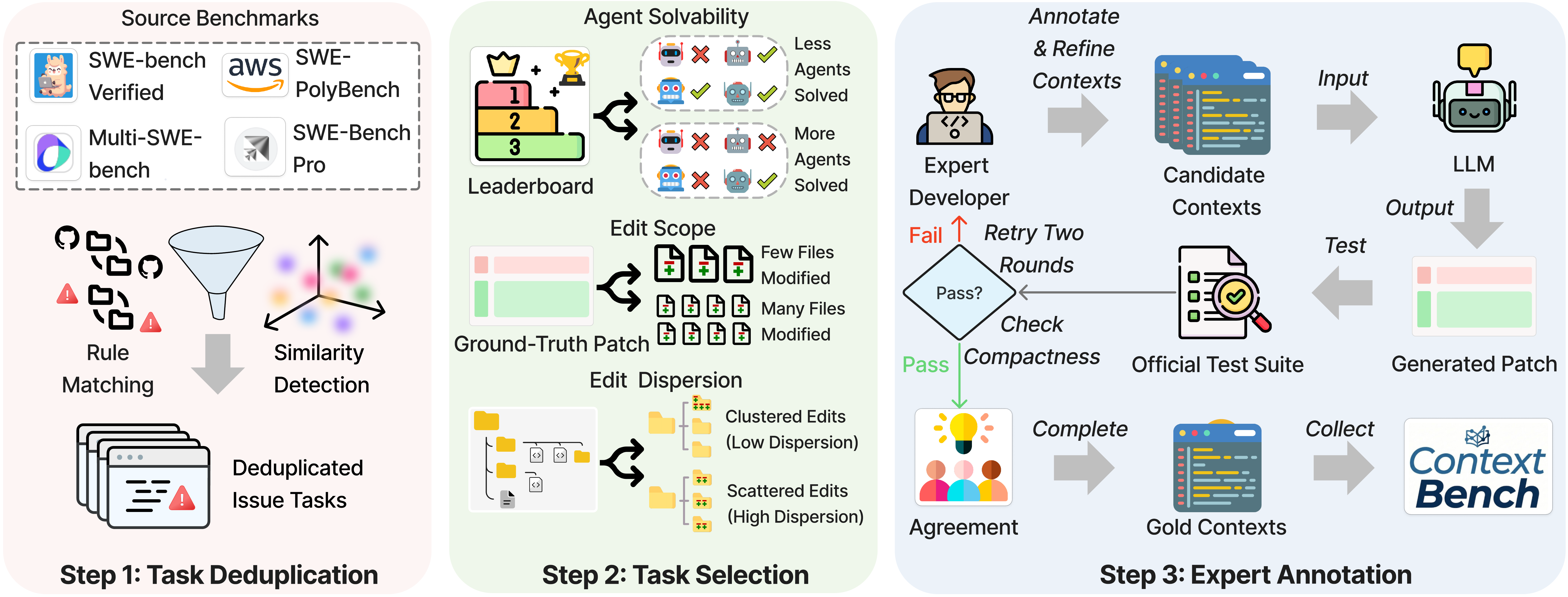}
         % \vspace{-2em}
         \caption{\textbf{An overview of the \benchmark construction pipeline.} \benchmark is curated through three key steps: \textbf{(1) Task Deduplication} removes exact and near-duplicate tasks from multiple issue resolution benchmarks using rule-based and embedding-based detection. \textbf{(2) Task Selection} identifies challenging tasks based on agent solvability and the scope and dispersion of edits in ground-truth patches. \textbf{(3) Expert Annotation} employs a human-in-the-loop procedure in which expert developers trace code dependencies to construct gold contexts, which are validated through LLM-based patch generation and inter-annotator agreement checks.}
	\label{fig_pipeline}
	\vspace{-1em}
\end{figure*}

As illustrated in~\autoref{fig_pipeline}, we present \benchmark, a benchmark for evaluating the agent's context retrieval ability in software engineering tasks, which is curated through three key steps:
\textbf{(1) Task Deduplication.}
We pool tasks from multiple issue resolution benchmarks and eliminate exact and near-duplicate tasks through both rule-based and embedding-based detection.
% We pool issue-resolution tasks from multiple existing benchmarks and curate a diverse candidate set via deduplication and difficulty-guided selection.
\textbf{(2) Task Selection.}
We select challenging tasks using three difficulty metrics, \textit{i.e.}, agent solvability, the scope and dispersion of edits in the ground-truth patch, prioritizing tasks that are complex or remain unsolved by existing agents.
\textbf{(3) Expert Annotation.}
For each selected task, expert developers iteratively annotate gold contexts in a human-in-the-loop process by tracing essential code dependencies from the ground-truth patches.
We then validate each annotated context by evaluating whether an LLM (\textit{e.g.}, GPT-5), conditioned solely on the context, can generate a patch that passes the official test suite or whether the context achieves high inter-annotation consistency.
% expert developers construct a compact, patch-driven gold context by tracing necessary dependencies from the ground-truth edits.
% We validate each gold context by conditioning a language model on it and requiring that at least one sampled patch passes the official test suite, ensuring sufficiency for issue resolution.
Based on this pipeline, we collect 1,136 task instances with verified gold contexts across 66 repositories and 8 programming languages.

% To verify the sufficiency of the constructed contexts, we provide them directly to an LLM (\textit{e.g.}, GPT-5) and evaluate whether the model can generate a patch that passes the official test suite under a multi-attempt evaluation setting.
% If the patch passes, annotators refine the context to ensure compactness; otherwise, annotation is repeated up to two additional rounds, and only tasks achieving high inter-annotation consistency (measured via context similarity) are retained in the final dataset.

% \subsection{Dataset Construction}
% \subsection{Step 1: Issue Instance Selection}
% \subsection{Step 1: Context Task Collection}
\subsection{Step 1: Task Deduplication}

We curate \benchmark by systematically sampling tasks from four widely used issue resolution benchmarks, \textit{i.e.}, SWE-bench Verified~\cite{jimenez2024swebench} (500 tasks), Multi-SWE-bench~\cite{zan2025multiswebench} (1632 tasks), SWE-PolyBench PB500~\cite{rashid2025swepolybench} (500 tasks), and SWE-bench Pro~\cite{deng2025swebenchpro} (1865 tasks), totaling 4,497 tasks.
These benchmarks offer high-quality task sources curated from real-world GitHub repositories and issues, with strong reproducibility guarantees and multilingual coverage across Python, Java, JavaScript, TypeScript, Go, Rust, C, and C++.

% duplicates -> 2000+ tasks
% diffcult threshold: 1136 tasks
% filter: exmaples
% cursor + GPT 5: annotate 1136 tasks
% passed/verified: 446 tasks (GPT-5), -> compact
% two-stage annotation, two persons, context similiarity, > 90 pass, compact
% verified: 500 (difficulty)
% filter

%Existing Benchmarks from SWE-Bench, SWE-Bench Pro, PolyBench.
%Say something selection criteria here, according to difficulty? How to determine difficulty? 
% We construct \benchmark by sampling issue instances from three established code-repair benchmarks: SWE-bench Verified~\cite{swebench-verified}, MultiSWE-bench~\cite{multiswebench}, and SWE-PolyBench~\cite{swepolybench}. Together, these sources provide a carefully curated pool of real-world GitHub repositories and issues with strong reproducibility support, spanning Python, Java, JavaScript, and other widely used languages.From this pool, we select 300 issue instances spanning both bug fixes and feature requests.

% \parabf{Task Deduplication.}
The source benchmarks originate from GitHub and may contain overlapping tasks across repositories.
To ensure uniqueness, we apply rule-based matching on task metadata (\textit{e.g.}, repository names and issue identifiers) to eliminate exact duplicates, yielding 3,981 tasks.
We further compute embedding-based semantic similarity between issue descriptions and remove near-duplicates above a predefined threshold (\textit{e.g.}, 0.9).
Finally, we manually inspect borderline cases flagged by the threshold to avoid false removals, resulting in 3,100 unique tasks.
This deduplication step ensures that each task in \benchmark corresponds to a distinct real-world issue, minimizing bias from redundant evaluations.
More implementation details are provided in~\cref{sec:appendix:detail_data_filtering}.
% 4497 -> 3981 -> 3.1k

% We conduct a sanity check to remove duplicate issues across source benchmarks. 
% We first identify exact duplicates using the repository name and issue identifier. 
% We then compute embedding-based semantic similarity between issue descriptions to detect near-duplicates, and manually review borderline cases flagged by the similarity threshold. 
% This procedure ensures that each instance in \benchmark corresponds to a distinct issue-repository pair, thereby mitigating bias from repeatedly evaluating identical or highly overlapping tasks.

% We perform a sanity check to remove duplicate issues across benchmarks based on repository name, issue identifier. we compute embedding-based semantic similarity between issue descriptions to identify near-duplicates, and manually review flagged borderline cases. This step ensures that each instance in \benchmark corresponds to a distinct issue-repository pair and avoids bias introduced by repeated evaluation of identical or highly overlapping tasks.

% \parabf{Complexity-Guided Selection.}
% \parabf{Task Selection.}
\subsection{Step 2: Task Selection}
To ensure that \benchmark evaluates challenging and non-trivial context retrieval scenarios, we rank candidate tasks using three complementary difficulty metrics, \textit{i.e.}, agent solvability, the scope and dispersion of edits in the ground-truth patch.

\parabf{Agent Solvability.} 
We automatically crawl results from the public leaderboards of the source benchmarks and use them as a coarse proxy for task difficulty.
For each task, we record the number of agents that successfully resolve it and prioritize tasks that remain unsolved or have been solved by only a few agents.

\parabf{Edit Scope.}
For each issue task, we analyze its ground-truth patch and record the number of modified files.
We prioritize tasks with larger edit scopes, as they typically require more extensive context retrieval and involve more complex reasoning patterns.
% Defined as the number of modified files in the ground-truth patch, this metric reflects the breadth of the search space required. 

\parabf{Edit Dispersion.} 
We measure the distribution of code edits across files and directories in the repository.
To quantify this, we compute the average structural distance between edited regions in the repository tree parsed by \texttt{Tree-Sitter}.
We prioritize tasks with highly dispersed edits that span multiple modules or distant files, since these cases involve broader contextual dependencies over scattered code regions.

% We measure how code edits are distributed across the repository by analyzing their spatial scattering over files and directories.
% To quantify this, we compute the average structural distance between edited regions in the repository tree constructed with the \texttt{Tree-Sitter} tool.

Based on these difficulty metrics, we select 1,500 candidate tasks.
We then manually review them to remove cases that appear challenging by metrics but are semantically trivial, such as large-scale variable renaming across multiple files or bulk formatting changes that do not affect program logic, resulting in a final set of 1,136 tasks.
More implementation details are provided in Appendix \ref{appendix-step2-agentsolvability}.
% 3.1k -> 1.5k
% 1.5k -> 1136

% This captures the spatial distribution of changes, specifically the span and scattering of modified code blocks across the repository. 
% We then select instances to ensure a diverse coverage of these dimensions, including both solvable and unsolved cases, ranging from localized single-file edits to complex, dispersed multi-file changes.

% To enable controlled evaluation across varying localization complexity, we rank candidate instances using three complementary difficulty signals. 
% (i) \textbf{Agent solvability}: we use agent outcomes reported by each source benchmark's leaderboard as a practical reference, recording for each instance whether it can be fully resolved by at least one submitted agent .
% (ii) \textbf{Edit scope}: the number of modified files in the ground-truth patch, which reflects how broadly an agent must search. 
% (iii) \textbf{Edit dispersion}: the overall span and dispersion of modified code blocks, which captures whether edits are concentrated or scattered across code regions. 
% We then select instances to cover a broad spectrum along these signals, including both solvable and unsolvable cases, and ranging from localized single-file edits to challenging multi-file, dispersed changes.

% \subsection{Step 2: Gold Context Annotation}
\subsection{Step 3: Expert Annotation}
\label{sec:expert_annotation}

\footnotetext{\url{https://github.com/langchain-ai/langchain/pull/4009}}

The annotation is conducted collaboratively by six authors of this paper and a group of engaged expert developers over four months.
To ensure quality and consistency, we adopt a \textit{human-in-the-loop} procedure where annotators iteratively construct and refine gold contexts for 1,136 selected tasks, validating them through LLM-based patch generation and inter-annotator agreement checks.
All annotators have over three years of software development experience working with large-scale codebases.
To ensure the annotators can proficiently mark the data, we provide detailed guidelines and examples that illustrate how to trace code dependencies, identify relevant artifacts, and record annotations in a standardized format.
We also specify clear criteria and task requirements to guide annotators throughout the process. Please refer to the details of our annotation process in~\cref{sec:appendix:a2}.

% We then validate each annotated context by evaluating whether an LLM (\textit{e.g.}, GPT-5), conditioned solely on the context, can generate a patch that passes the official test suite or whether the context achieves high inter-annotation consistency.

% For each selected task, expert developers iteratively annotate and refine the required contexts in a human-in-the-loop process, ensuring that the resulting gold contexts are compact and sufficient for issue resolution.

% \zychu{TODO: Formalize what ``gold context'' means. Rationale: clarify that gold context is a compact but sufficient set (not guaranteed minimal) so readers can correctly interpret precision/F1; provide the formal definition and scope in Appendix~\ref{sec:appendix:gold_context_definition}.}

\parabf{Context Annotation.}
The annotation process begins with the ground-truth patch for each issue task.
Annotators follow the provided guidelines to trace essential code dependencies and semantically related artifacts originating from the modified regions.
For each edited location, they inspect its function and class invocations, inheritance relations, and related control-flow and data-flow paths.
Annotators also review the surrounding code within the same file or module to identify components that are semantically relevant to the modification.
Through this tracing process, annotators are instructed to ensure that the recorded contexts are sufficient for issue resolution while remaining as compact as possible by excluding redundant or irrelevant code. 

% For each selected issue, we construct a verified gold context that is as compact as possible while remaining sufficient for issue resolution. Our annotation is primarily \textbf{patch-driven}: starting from the ground-truth patch, we identify the edited code regions and trace the necessary dependencies to form a candidate gold context.

\begin{figure*}[!t]
	\centering
	\includegraphics[width=0.98\linewidth]{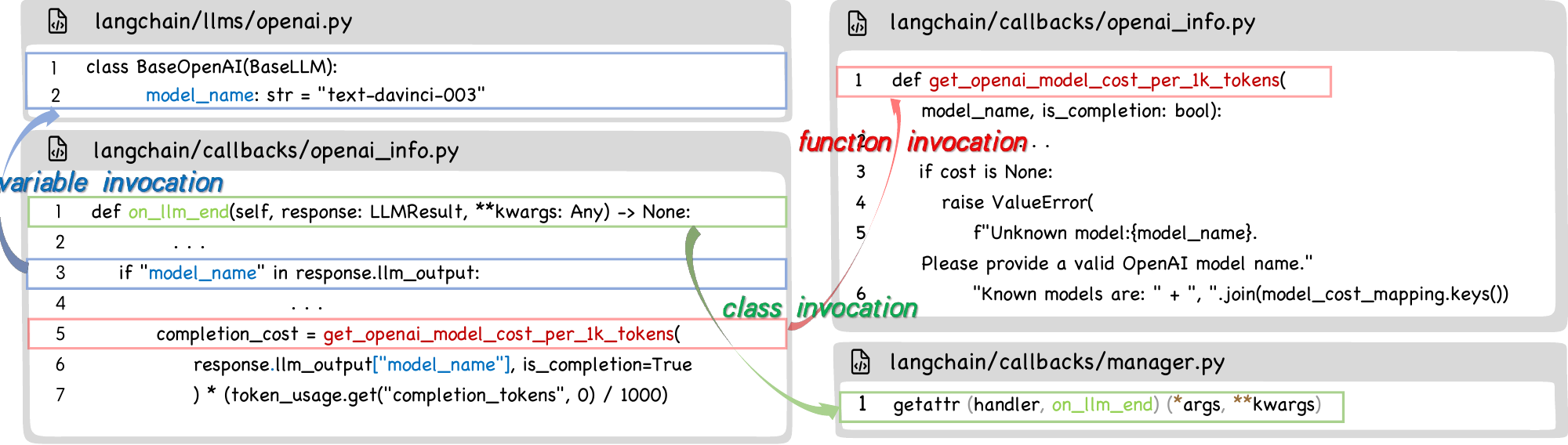}
         % \vspace{-2em}
         \caption{\textbf{An illustrative example of the human-verified gold context.} This example is derived from the issue task associated with Pull Request \texttt{\#4009} in the \texttt{langchain-ai/langchain} repository. 
         The blue, red, and green arrows represent the manually annotated contextual dependencies associated with variable, function, and class invocations, respectively.
         % The blue arrow highlights the contextual link of a variable invocation, the red arrow indicates a function invocation, and the green arrow represents a class invocation.
         }
	\label{fig_example}
	\vspace{-1em}
\end{figure*}

\parabf{Context Verification.}
To validate whether each annotated context is sufficient to resolve its corresponding issue, we utilize a state-of-the-art LLM (\textit{e.g.}, GPT-5) as an evaluator.
For each task, the model is conditioned solely on the annotated context and prompted to generate multiple candidate patches (\textit{e.g.}, 5 attempts).
We then consider an annotated context sufficient if at least one generated patch successfully passes the official test suite, including both fail-to-pass and pass-to-pass test cases.
% \zychu{TODO: Briefly justify why a single verifier (GPT-5) is sufficient for our feasibility-oriented sufficiency check; defer full rationale to Appendix~\ref{sec:appendix:detail_context_verification}.}
% We use GPT-5 as the single verifier for context sufficiency; detailed rationale and protocol are in Appendix~\ref{sec:appendix:detail_context_verification}.
Detailed rationale and protocol are provided in~\cref{sec:appendix:detail_context_verification}.

% To verify the sufficiency of each annotated context, we provide it to an LLM (\textit{e.g.}, GPT-5) and evaluate whether the model can generate a patch that passes the official test suite under a multi-attempt evaluation setting.
% We construct an initial candidate context by anchoring on the modified code spans in the ground-truth patch. Annotators then trace outward to include the relevant files, functions, variables, and non-local dependencies required to understand and implement the change. The resulting set of code elements constitutes the candidate gold context for the issue.

\parabf{Context Refinement.}
When an annotated context passes verification, we assign it to a different annotator, distinct from the one who performed the initial annotation, to conduct compactness checking.
This secondary annotator reviews the verified context to identify and remove any code segment that is redundant or irrelevant to issue resolution.
Subsequently, the two annotators jointly review the updated context, resolve any disagreement, and finalize a consensus version of the gold context.

If an annotated context fails verification, we conduct two additional rounds of annotation and verification, each handled by a different annotator. 
In each round, the new annotator identifies the context by independently tracing code dependencies and then validates it through the LLM-based verification procedure described earlier.
This iterative process accounts for the possibility that some tasks are inherently difficult for LLMs to solve, not necessarily due to incomplete context annotation.
After the two additional rounds, all three annotators jointly inspect the resulting contexts, reconcile differences, and refine them into a final context version.

Consequently, the annotation process produces a gold context represented by a compact set of code regions that are essential for resolving the issue and generating the corresponding patch, as illustrated in \autoref{fig_example}.
This collaborative refinement ensures that each gold context in \benchmark is both semantically sufficient for issue resolution and as concise as possible.
% \zychu{TODO: Statistics}

% If the patch passes, annotators refine the context to ensure compactness; otherwise, annotation is repeated up to two additional rounds, and only tasks achieving high inter-annotation consistency (measured via context similarity) are retained in the final dataset.
% To ensure that constructed contexts are sufficient for issue resolution, we introduce a verification stage shared by both trajectory reused and human annotated contexts. Specifically, we provide the gold context directly to a language model and sample multiple candidate patches. If at least one generated patch passes the corresponding test suite, including fail to pass and pass to pass cases, we consider the context sufficient. Contexts that fail verification are refined or discarded. This procedure ensures that \benchmark contains only validated gold contexts that can, in principle, lead to correct solutions.

% \parabf{Benchmark Statistics.}
\subsection{\benchmark Statistics}
\begin{figure*}[!t]
    \vspace{-1em}
	\centering
	\includegraphics[width=0.98\linewidth]{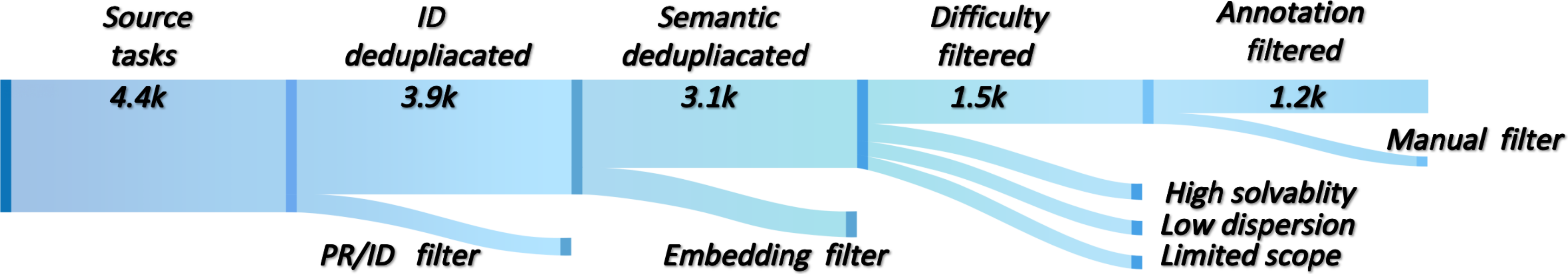}
         % \vspace{-2em}
         \caption{\textbf{The data flow during the \benchmark construction pipeline.} We construct \benchmark step by step, starting from task deduplication and selection to gold context annotation.}
	\label{fig_data_flow}
	\vspace{-1em}
\end{figure*}

% After issue selection, gold context annotation, and verification, we curate \benchmark as a unified dataset of issue resolution instances with verified gold contexts. 
% The dataset consists of both resolved and unresolved issues drawn from existing benchmarks, enabling evaluation of context retrieval behavior under diverse success conditions. 
\autoref{fig_data_flow} illustrates the data filtering pipeline, showing how tasks are processed through deduplication, selection, and expert annotation.
This pipeline produces \benchmark, a repository-level benchmark featuring human-verified gold contexts. 
As summarized in~\autoref{tab:dataset_overview}, \benchmark comprises 66 repositories and 1,136 issue tasks across eight programming languages, with context annotations provided at the file, block, and line levels, encompassing 4,548 files, 23,116 blocks, and 522,115 lines of code.
These gold contexts enable systematic evaluation of intermediate context retrieval beyond final task resolution rates.
To facilitate efficient evaluation, we further construct a Lite subset of 500 tasks, selected based on coding agent solvability and the scope and dispersion of edits in the ground-truth patch. 
% Table~\ref{tab:dataset_overview} provides an overview of \benchmark, a repository-level benchmark with human-annotated gold contexts. 

% \begin{table}[!t]
% % \fontsize{7pt}{12pt}\selectfont
% % \setlength{\tabcolsep}{4pt} 
% \begin{minipage}[t]{\linewidth}
% \caption{\textbf{Statistics of data in \benchmark.} We construct \benchmark step by step, from XXXX, XXXXX, and XXXX to XXX. \zychu{This is a PLACEHOLDER, please fill it in. You can refer to the papers ``CODESYNC: Synchronizing Large Language Models with Dynamic Code Evolution at Scale'' and ``MLLM-as-a-Judge: Assessing Multimodal LLM-as-a-Judge with Vision-Language Benchmark''}}
% \centering
% \begin{tabular}{cccccc}
%     \toprule[1.5pt]
%     \textbf{Step} & \textbf{Setting} & \textbf{Input} & \textbf{Num.} & \textbf{Output} & \textbf{Num.} \\
%     \midrule
%     1 & - & - & - & - & - \\
%     \midrule
%     2 & - & - & - & - & - \\
%     \midrule
%     3 & - & - & - & - & - \\
%     \midrule
%     \makecell{\textsc{Context} \\ \textsc{Bench}} & - & - & - & - & - \\
%     \bottomrule[1.5pt]
% \end{tabular}
% \label{tab_dataset_statistics}
% \end{minipage}
% \vspace{-1em}
% \end{table}

% \subsection{\benchmark Statistics}
% \label{sec:curation}

% \paragraph{Context Coverage}

\input{tables/context-bench-overview-table}

% \subsection{\benchmark Evaluation Framework}
% \label{sec:metrics}

\subsection{\benchmark Evaluation Protocol}

% \subsection{Evaluation Pipeline}
To evaluate the context retrieval ability of coding agents, we instrument their execution to trace the code regions they inspect during issue resolution and subsequently compare their retrieved contexts with human-verified gold contexts.

\parabf{Agent Context Tracing.}
For each issue task, we instruct the agent to perform the full resolution process while automatically recording every accessed code segment, including the corresponding file path, line range, and content.
We then apply regular expressions to extract these regions and standardize them into structured \texttt{<PATCH\_CONTEXT>} blocks, which serve as intermediate checkpoints representing the agent’s context snapshots throughout its execution trajectory.
Moreover, we require the coding agent to explicitly declare, before submitting its final patch, the code context it considers essential for generating the patch.
This final declared context may not completely overlap with the intermediate checkpoint contexts, as it reflects the agent’s own judgment of what information is most relevant to the patch.
More implementation details are provided in~\cref{sec:appendix:detail_context_tracing}.

\parabf{Agent Context Evaluation.}
We then parse the entire repository with the \texttt{tree-sitter} tool to construct a unified structural coordinate system, onto which both the gold and agent contexts are mapped for comparison.
The two context sets are aligned across three granularities:
(i) the file level, by matching file paths;
(ii) the block level, by aligning \textit{Abstract Syntax Tree} (AST) nodes at \texttt{definition} level, including classes, functions, interfaces, and trait declarations (detailed in~\cref{sec:appendix:block_nodes}); and
(iii) the line level, by comparing line ranges using their byte offsets within files.
This structured alignment enables fine-grained comparison between the gold and agent contexts.
Based on this alignment, we evaluate recall, precision, F1, and process-level context dynamics (\textit{e.g.}, efficiency, redundancy, and usage) at the file, block, and line levels, with detailed definitions provided in~\cref{sec:appendix:metrics}.

% \zychu{TODO: clarify block-level alignment: we evaluate definition-level symbols (functions/methods/classes/interfaces/traits), not arbitrary AST nodes. Add an explicit pointer to Appendix~\ref{sec:appendix:block_nodes} and ensure the node-type table matches the implementation.}

% Replace the block-level sentence with the following:
% (ii) the block level, by aligning \emph{definition-level} AST nodes (e.g., function/method/class/interface/trait declarations) extracted by \texttt{tree-sitter}; see Appendix~\ref{sec:appendix:block_nodes}; and

% We then utilize the \texttt{tree-sitter} tool to parse the corresponding repository and map both gold and agent contexts onto a shared coordinate system that compare their file paths, and compare AST blocks (e.g., classes and functions), and line ranges using byte offset

\input{tables/experiment1}

% \section{Experiment and Results}
\section{Can Coding Agents Retrieve Effective Contexts to Resolve Issues?}
\label{sec:experiments}

To evaluate coding agents' capability in retrieving effective contexts for software engineering tasks, we investigate the following \textit{Research Questions} (RQs):
\begin{itemize}[leftmargin=4mm, itemsep=0.05mm]
    \item \textbf{RQ1: Benchmarking Coding Agents.} 
        \textit{How effectively do coding agents retrieve contexts from large-scale repositories when resolving issues using a state-of-the-art LLM (\textit{i.e.}, GPT-5)?}
    \item \textbf{RQ2: Benchmarking Large Language Models.}
        \textit{How effectively do LLMs retrieve code contexts for issue resolution when equipped with a standard coding agent framework (\textit{i.e.}, mini-SWE-agent)?}
    \item \textbf{RQ3: Analysis of Context Retrieval Patterns.} 
        \textit{What characteristic patterns emerge when LLM agents retrieve code contexts?} 
    \item \textbf{RQ4: Analysis of Context Retrieval Dynamics.} 
        \textit{What dynamics emerge in LLM agents’ context retrieval processes during task execution?} 
    % \item \textbf{RQ5: Analysis of Agent Failures.} 
    %     \textit{What other factors contribute to failures beyond insufficient code context?} 
     \item \textbf{RQ5: Analysis of Gold Context Robustness.} 
        \textit{How robust are the human-verified gold contexts?} 
        
\end{itemize}

% \begin{tcolorbox}[colback=lightgray!10, colframe=black, title={Finding 1}]
% \textbf{Backbones exhibit distinct retrieval preferences.} Different LLMs trace different coverage--precision trade-offs under the same agent scaffold.
% \end{tcolorbox}
% \vspace{-6pt}
% \noindent

\subsection{RQ1: Benchmarking Coding Agents}

\autoref{tab:contextbench_main} benchmarks five coding agents that are open-sourced, actively maintained, and representative of major paradigms in context retrieval.
Specifically, we assess a standard baseline that retrieves context through simple bash commands (\textit{i.e.}, min-SWE-agent~\cite{yang2024sweagent}) and four state-of-the-art scaffolds, including embedding-based semantic retrieval (\textit{i.e.}, Agentless~\cite{xia2024agentless}), file-system navigation via specialized interfaces (\textit{i.e.}, SWE-agent~\cite{yang2024sweagent}, OpenHands~\cite{wang2025openhands}), and graph-based repository retrieval (\textit{i.e.}, Prometheus~\cite{chen2025prometheus}).
All agents are evaluated with an advanced LLM backbone (\textit{i.e.}, GPT-5~\cite{openai2025gpt5}).
% All experimental results for the five agents are based on the same foundation model (GPT-5~\cite{openai2025gpt5}) and use the same evaluation metrics as in RQ1. 
Detailed experiment settings are listed in~\cref{sec:appendix:a1.2}.

The results reveal distinct context retrieval behaviors across coding agents, driven by differences in their tool interfaces and control policies.
Some agents (\textit{e.g.}, Prometheus) favor broad exploration and retrieve larger contexts, achieving higher recall but lower precision, whereas others adopt more conservative strategies.
Yet, despite differences in retrieval strategies, their overall context metrics often fall below those of the baseline mini-SWE-agent.
This finding suggests that current agent frameworks may be \textit{over-engineered} for context retrieval. 
In practice, simple iterative search and inspection through basic shell commands can already recover a substantial portion of the necessary context, whereas more complex orchestration introduces additional overhead and may even degrade retrieval performance.

\begin{summary}
\textbf{Answer to RQ1:}
Sophisticated agent scaffolding does not necessarily improve context retrieval performance, revealing potential \textit{over-engineering} in current designs, which echoes \textbf{``The Bitter Lesson''} of AI research.
% \textbf{Finding 2~\raisebox{-0.2ex}{\includegraphics[height=1em]{figures/bulb.pdf}}:}
% Agent designs change costs more than retrieval quality. Different retrieval strategies yield similar F1, but substantially different token and context costs.
% Different coding agents use different context retrieval strategies, yet they yield only marginal gains over a minimal foundation-model agent baseline.
\end{summary}

\subsection{RQ2: Benchmarking Large Language Models}

\autoref{tab:backbone_compare} benchmarks four state-of-the-art LLMs, including GPT-5~\cite{openai2025gpt5}, Claude Sonnet 4.5~\cite{anthropic2025claudesonnet45}, Gemini 2.5 Pro~\cite{google2025gemini25}, and Devstral 2~\cite{mistral2025devstral2}, in retrieving relevant code contexts from large-scale codebases when resolving issues using the standard coding agent scaffold (\textit{i.e.}, mini-SWE-agent~\cite{yang2024sweagent}).
Detailed experiment settings are listed in~\cref{sec:appendix:a1.2}.

% Table~\ref{tab:backbone_compare} compares four \emph{backbone models} under the same mini SWE-agent scaffold, isolating the impact of the underlying LLM on retrieval quality and efficiency. We evaluate code context coverage, precision and F1 score at the file, symbol, and span levels, together with a process-oriented AUC coverage metric. Finally, the last column reports the pass@1 issue resolution rate. 

The results suggest that state-of-the-art LLMs still face challenges in retrieving effective context during issue resolution.
For instance, their block-level F1 scores fall below 0.45, while line-level F1 scores remain below 0.35.
% We make the following observations in these experiments.
Moreover, higher recall does not necessarily indicate better context retrieval performance, as a clear trade-off exists between recall and precision. 
% There is a clear trade-off between recall and precision in context retrieval. 
Some LLMs tend to expand retrieved contexts aggressively to achieve higher recall, while introducing excessive irrelevant content that results in lower precision.
For example, GPT-5 achieves higher recall at both the block and line levels but sacrifices precision, leading to lower overall F1 and, consequently, reduced issue resolution performance compared to Claude Sonnet 4.5.
% Second, \benchmark remains sufficiently challenging. Claude Sonnet 4.5 resolves only 53.0\% of issues, whereas under the same mini SWE-agent setting, the same model resolves 74.4\% of issues on SWE-Bench Verified.

% This relationship supports the central premise of \benchmark: context retrieval quality is not merely a diagnostic side metric, but a meaningful predictor of end-to-end performance. 
% At the same time, the recall-precision tension implies that improving success via higher recall must be coupled with better evidence selection; otherwise, excessive context can reduce precision and introduce noise that undermines localization.

We also observe a positive correlation between context recall and downstream task success.
LLM agents that cover a larger portion of gold contexts tend to achieve higher Pass@1 scores.
This finding reinforces the central premise of \benchmark: context retrieval quality is not merely a diagnostic auxiliary metric but a meaningful predictor of end-to-end performance.
However, the observed recall-precision trade-off suggests that gains from broader retrieval must be balanced with more accurate evidence selection; otherwise, excessive context can introduce noise and impair the agent’s reasoning and issue resolution performance.

% Table~\ref{tab:backbone_compare} shows that backbone choice shifts not only absolute retrieval quality but also the \emph{coverage--redundancy frontier}. 
% In particular, some backbones are more inclined to expand context aggressively, resulting in higher coverage but lower precision. For example, GPT-5 tends to retrieve and inspect more context, which raises coverage yet dilutes precision; consequently, the overall F1 does not necessarily improve. In contrast, Claude exhibits a more balanced profile, yielding more stable trade-offs across granularities. These consistent differences indicate that retrieval behavior is shaped by the backbone’s implicit preference for \emph{evidence breadth} versus \emph{evidence selectivity}, affecting both redundancy and the quality of the final evidence set.

\begin{summary}
\textbf{Answer to RQ2:}
State-of-the-art LLMs struggle with effective code context retrieval, often covering relevant information at the cost of introducing substantial noise and sacrificing retrieval precision.
% \textbf{Finding 1~\raisebox{-0.2ex}{\includegraphics[height=1em]{figures/bulb.pdf}}:}
% Backbones exhibit distinct retrieval preferences. Different LLMs trace different coverage--precision trade-offs under the same agent scaffold.
\end{summary}

\input{tables/model_backbone}

\subsection{RQ3: Analysis of Context Retrieval Patterns}

\autoref{tab:model-strategies} compares the context retrieval patterns across LLM agents on \benchmark, measured by the number of retrieval steps per instance, lines retrieved per step, total lines retrieved per instance, and cost per instance.

We can observe that LLMs adopt different context retrieval strategies, trading off retrieval steps against retrieval granularity. GPT-5 retrieves context in the fewest rounds (5.87), with substantially more lines per step (119.29 lines of code on average), whereas Devstral~2 performs the most retrieval rounds (22.16) while retrieving much smaller contexts per step (11.98 lines of code, approximately 10\% of GPT-5).
Notably, Claude Sonnet~4.5 adopts a more balanced strategy, combining moderate retrieval rounds with moderate context size. Recall in \autoref{tab:backbone_compare}, this balanced behavior aligns with its strong performance, achieving the best line-level F1 score and Pass@1 resolution rate.

The last column of \autoref{tab:model-strategies} reports the average cost per instance for context retrieval.
Devstral 2 is the most expensive due to more retrieval rounds that produce costly output tokens. This suggests that reducing the number of queries is an effective way to lower LLM cost.

\begin{summary}
\textbf{Answer to RQ3:}
LLMs differ in how they balance retrieval rounds and context granularity, and models with more balanced strategies achieve better line-level retrieval quality and end-to-end resolution performance.
\end{summary}

%%% talk about the cost.

\subsection{RQ4: Analysis of Context Retrieval Dynamics}
% \paragraph{Retrieval Dynamics Along Trajectories.}

% \begin{table}[t]
% \centering
% \small
% \setlength{\tabcolsep}{5pt}
% \caption{\textbf{RQ4.}}
% \begin{tabular}{lcc}
% \toprule
% Agent &
% AUC-Cov &
% Redun \\
% \midrule
% Agentless  & 0.056 & 0.000 \\
% Prometheus & 0.598 & 0.422 \\
% OpenHands  & --    & --   \\
% SWE-Agent  & 0.563 & 0.094 \\
% \bottomrule
% \end{tabular}
% \vspace{2pt}

% \label{tab_rq4_agents}
% \end{table}

Beyond evaluating the final aggregated contexts submitted before patch generation, we further analyze how coding agents retrieve context step by step during execution, using the metrics detailed in~\cref{sec:appendix:metrics}.
As shown in \autoref{tab_rq4_models}, LLM agents exhibit different context retrieval dynamics.
Among them, Claude Sonnet 4.5 demonstrates a particularly unique behavior: it achieves the highest efficiency, capturing relevant gold contexts earlier than other models, but at the cost of substantial redundancy.
This behavior may stem from its exploration strategy, in which the model repeatedly revisits previously accessed files to maintain coverage of potentially relevant code regions, leading to redundant retrieval of overlapping content.

We also observe a consistent context usage drop across all evaluated agents, indicating that a considerable portion of the retrieved context during execution is ultimately unused in the final patch generation.
Consequently, many gold contexts successfully retrieved at intermediate steps are not incorporated into the final reasoning process, leading to issue resolution failures despite retrieval success.
Notably, Gemini 2.5 Pro and Devstral 2 exhibit the most severe context loss, which likely explains their relatively low issue resolution rates.

\begin{summary}
\textbf{Answer to RQ4:}
LLM agents exhibit diverse context retrieval dynamics, revealing inherent trade-offs between efficiency and redundancy.
Moreover, the significant drop in context usage hinders effective issue resolution.
% Consolidation drops relevant evidence. Agents often inspect gold-relevant regions, yet fail to retain all of them in $\tilde{A}$.
\end{summary}
% \begin{summary}
% \textbf{Answer to RQ4-B:}
% Late-stage retrieval wastes budget and amplifies noise. Coverage often saturates early, while continued retrieval inflates redundancy and lowers precision.
% \end{summary}

\input{tables/rq3}

% \subsection{RQ5: Investigating Agent Failures}
% \label{sec:failures}

% To understand why agents with high exploration coverage can still fail end-to-end, we conduct a focused analysis on a 397-instance diagnostic subset.
% We examine: (i) the mismatch between cumulative explored context $A^{(t)}$ and final patch context $\tilde{A}$ (evidence drop at consolidation time); (ii) inefficiency patterns captured by AUC-Cov and redundancy (late-stage over-retrieval); and (iii) how different retrieval interfaces (ranking-based vs.\ interactive exploration) affect the recall--precision trade-off and localization stability.

% (Optional) prevent floats from crossing into Related Work if you keep adding figures/tables here
% \FloatBarrier

\subsection{RQ5: Analysis of Gold Context Robustness}

% \paragraph{Benchmarking Stability.}
% We further study stability under \emph{solution multiplicity} by comparing contexts induced by multiple distinct test-passing patches versus a single patch for the same issue.
% This disentangles instability caused by alternative valid solutions from errors introduced by agent retrieval and consolidation, and supports the robustness of \benchmark's gold contexts as a supervisory signal for retrieval-centric evaluation.

% To assess robustness under \emph{solution multiplicity}, we analyze how gold contexts vary across multiple distinct \emph{test-passing} patches for the same issue.

% \parabf{Alternative-Patch Consistency.}
% For 82 instances, we collect $2$--$3$ different test-passing patches per instance provided by existing benchmark variants, and derive a patch-conditioned gold context for each patch using the same patch-driven tracing procedure. 

Our gold contexts are human-annotated based on a gold patch. However, in practice, a single issue may admit multiple semantically equivalent but syntactically different patches, raising the question of whether the collected gold context is reliable under solution multiplicity.

To answer this question, we analyze how gold contexts vary across multiple distinct test-passing patches for the same issue. Specifically, we conduct a case study on 82 tasks from \benchmark, each with two semantically equivalent patches, for which we derive corresponding patch-conditioned gold contexts using the same context tracing procedure.
% For each task, we collect two semantically equivalent patches and derive a patch-conditioned gold context for each patch using the same patch-driven context tracing procedure.
We quantify consistency using the Jaccard similarity between each pair of contexts, detailed in~\cref{sec:appendix:metrics}. For each task, we report the average similarity across all context pairs.
Across all evaluated instances, the average Jaccard similarity is 0.9518 (average distance 0.0482), indicating that our gold contexts are highly consistent across alternative valid patches.
% \zychu{TODO: heat map}

\begin{summary}
\textbf{Answer to RQ5:}
The high consistency of gold contexts across semantically equivalent patches supports their reliability for context retrieval evaluation.
\end{summary}

\begin{table*}[!t]
\centering
\footnotesize
\caption{\textbf{Context retrieval dynamics of different LLM agents.} LLM agents show diverse retrieval dynamics, exposing trade-offs between efficiency and redundancy. In addition, the significant loss of context usage impedes effective issue resolution.
}
\begin{tabular}{lccc}
\toprule[1.5pt]
\multirow{1}{*}{\textbf{LLM}} &
\textbf{Efficiency$\uparrow$} &
\textbf{Redundancy$\downarrow$} & 
\textbf{Usage Drop$\downarrow$} \\
\midrule
GPT-5              & 0.591 & \cellcolor{tablelightpurple}\textbf{0.487} & \cellcolor{tablelightpurple}\textbf{0.179}\\
Claude Sonnet 4.5  & \cellcolor{tablelightpurple}\textbf{0.658} & 0.708 & 0.196\\
Gemini 2.5 Pro     & 0.529 & 0.558 & 0.431\\
Devstral 2         & 0.616 & 0.672 & 0.435\\
\bottomrule[1.5pt]
\end{tabular}
% \vspace{2pt}
\label{tab_rq4_models}
\end{table*}

\section{Related Work}

\parabf{Repository-Level Code Evaluation Benchmarks.}
Repository-level benchmarks ground evaluation in real-world codebases and executable test suites, making them the de facto standard for assessing end-to-end issue resolution. SWE-bench and its variants (Lite, Verified, multilingual extensions \cite{zan2025multiswebenchmultilingualbenchmarkissue}, long-horizon tasks \cite{deng2025swebenchproaiagents}, live environments \cite{zhang2025swebenchgoeslive}, goal-oriented settings \cite{yang2025codeclashbenchmarkinggoalorientedsoftware}, and freelance scenarios \cite{miserendino2025swelancerfrontierllmsearn}) evaluate whether a system can produce a test-passing patch under realistic repository settings \cite{yang2024sweagent,jimenez2024swebench}, while complementary benchmarks such as RepoBench \cite{liu2023repobench} and SWE-PolyBench \cite{rashid2025swepolybench} emphasize cross-file completion or structure-aware analysis. Despite their realism, these benchmarks are predominantly \emph{outcome-driven}: they largely measure final task success, offering limited visibility into whether agents succeed by identifying the right information or by compensating through extensive exploration \cite{chen2025prometheus}. 
Our work complements outcome-only evaluation by (i) introducing \emph{verified gold contexts} and (ii) enabling dynamic, process-level evaluation of context retrieval beyond final outcomes.

\parabf{Coding Agents.}
LLM-based coding agents address repository-level issue resolution through either pipeline-based or agentic designs. Pipeline-based systems decompose tasks into fixed stages with controlled context selection \cite{xia2024agentless, zhang2024autocoderover, ruan2024specrover, gao2025trae, yang2025lingxi}, agentic systems adopt open-ended interaction loops in which agents dynamically search, inspect, and modify codebases via tool calls \cite{yang2024sweagent, wang2025openhands, xia2025live, chen2025swe, chen2025prometheus}. 
Despite architectural diversity, context retrieval remains a shared bottleneck. Existing benchmarks primarily evaluate end-to-end success, making it difficult to compare how agents retrieve and manage code context. \benchmark fills this gap by providing a retrieval-centric benchmark that directly evaluates context acquisition behavior, independent of final patch correctness.

\parabf{Trajectory Analysis and Failure Diagnosis.}
Recent work analyzes agent trajectories to diagnose failure modes via manual inspection, error taxonomies, or large-scale statistics over execution logs~\cite{zhu2025agentdebug, liu2025empirical, chen2025errors, bouzenia2025understanding, cemri2025multi, pathak2025detecting}. Such analyses have improved understanding of where agents fail (e.g., reasoning deadlocks, tool misuse, runtime exceptions), and some studies report that context-related errors are frequent in repository-level settings. However, trajectory analysis is often labor-intensive, difficult to standardize across frameworks, and typically provides \emph{post hoc} explanations rather than instance-level supervision. \benchmark addresses this gap by offering a scalable, standardized evaluation layer: verified gold contexts and automatic metrics that quantify retrieval effectiveness without requiring manual reading of long trajectories, while still enabling connections between intermediate retrieval behavior and end-to-end outcomes.

\section{Conclusion}
We introduced \benchmark, a benchmark for process-level evaluation of context retrieval in LLM-based coding agents. By augmenting existing issue-resolution benchmarks with human-annotated gold contexts and retrieval-centric metrics, \benchmark enables analysis beyond end-to-end success rates. Our evaluation across four frontier LLMs and five coding agents reveals limited benefits from complex agent scaffolding, consistent recall-over-precision retrieval behavior, and notable gaps between retrieved and utilized context. These results highlight the need for process-oriented evaluation and suggest intermediate context signals as a promising direction for improving LLM-based software engineering systems.

This work introduces a new process-level evaluation paradigm for LLM-based coding agents, enabling the community to move beyond coarse-grained end-to-end metrics toward a deeper understanding of agentic reasoning and context utilization. By providing verified gold contexts and automated retrieval metrics, \benchmark facilitates transparent, reproducible, and fine-grained assessment of how coding agents interact with large-scale codebases.
We expect this benchmark to support the development of more reliable and interpretable software engineering agents, improving their robustness in real-world development workflows. In the broader context, this work contributes to responsible AI research by promoting rigorous evaluation practices, reducing overfitting to leaderboard-style benchmarks, and fostering more trustworthy integration of AI systems in software engineering and related domains.

% \section*{Acknowledgments}
% Add your acknowledgments here.
% \clearpage

% =========================
% REFERENCES
% =========================
\bibliographystyle{plain}
\bibliography{custom}

\clearpage
\appendix

\label{sec:appendix}

\section{Details on Data Filtering}
\label{sec:appendix:detail_data_filtering}

This subsection provides a detailed description of the data filtering pipeline used to construct the final evaluation task set.
The filtering process is designed to progressively remove redundancy, low-informative tasks, and annotation artifacts, while preserving task diversity and evaluation difficulty.

\paragraph{Initial task pool.}
We begin with an initial pool of approximately 4.4k source tasks aggregated from existing software engineering benchmarks, including SWE-bench Verified, SWE-PolyBench, and other related curated datasets.
Although these tasks originate from real-world issue and pull request scenarios, our work does not directly crawl raw repositories; instead, it consolidates and refines tasks from established benchmarks.
At this stage, the task set intentionally serves as a broad superset and contains substantial redundancy and variability in task characteristics.

\paragraph{ID-based deduplication.}
As a first step, we apply an ID-level deduplication procedure that removes tasks with identical or invalid issue/PR identifiers.
This step eliminates exact duplicates introduced during dataset aggregation and ensures that each remaining task corresponds to a unique identifier.
After this filtering, 3.9k tasks remain.

\paragraph{Semantic deduplication.}
Next, we perform semantic-level deduplication using an embedding-based filtering strategy.
Each task is encoded into a semantic vector representation using a pretrained embedding model, and pairwise cosine similarity is computed between task embeddings.
Tasks whose cosine similarity exceeds a predefined threshold of 0.90 are considered semantically redundant and are filtered out, even if their identifiers differ.
This step removes near-duplicate tasks that differ only in superficial wording while preserving semantically distinct problem instances.
After semantic deduplication, the task set is reduced to 3.1k tasks.

\section{Details on Agent Solvability}
\label{appendix-step2-agentsolvability}
\paragraph{Difficulty-based filtering.}
The third stage applies difficulty-based filtering and accounts for the largest reduction in task count.
Specifically, we exclude tasks that fall into one of the following categories:
(i) \emph{high solvability}, where the task is trivially solvable by most models and provides limited evaluative value;
(ii) \emph{low dispersion}, where model performance exhibits minimal variance across different capability levels;
and (iii) \emph{limited scope}, where the task can be resolved using only minimal and easily obtainable context.
Concretely, tasks classified as limited scope typically involve fewer than four distinct code hunks in the annotated gold context provided to the agent, and therefore do not sufficiently exercise context retrieval or multi-step reasoning capabilities.
This step is critical for ensuring that the benchmark meaningfully differentiates model behaviors.
After difficulty-based filtering, 1.5k tasks remain.

\paragraph{Annotation filtering.}
Finally, we apply a manual annotation filter to ensure annotation quality and consistency.
Tasks are retained only if they are associated with high-quality gold context annotations that are complete, precise, and suitable for structured analysis.
This manual review step removes residual noise and annotation artifacts that are difficult to detect automatically.
The final benchmark consists of 1.2k tasks.

\clearpage
\section{Implementation Details on Agent Context Tracing}
\label{sec:appendix:detail_context_tracing}

We describe our context extraction mechanism and the adaptations applied across multiple agents. 
For clarity, we divide this section into two parts: A1.1 focuses on model variations within the mini-SWE-agent; 
A1.2 focuses on cross-agent differences for the same model under varying prompt designs and context extraction protocols.

%==================== A1.1 ====================
\subsection{Model Variations within mini-SWE-Agent}
\label{sec:appendix:a1.1}

\subsubsection{Context Extraction Architecture}

In this subsection, we analyze how different models behave under the mini-SWE-agent context extraction protocol. 
We extend the default agent architecture with a \texttt{ContextAwareAgent} class to enable structured and standardized context collection.

To enforce strict structure and machine-readability, prompts are augmented with predefined templates, and the \texttt{<PATCH\_CONTEXT>} block is extracted using a verification function with regular expressions. 
This setup ensures consistency, reproducibility, and automated verification within mini-SWE-agent workflows.

\subsubsection{Two-Stage Submission Protocol}

The submission workflow follows a two-stage protocol. In the first stage, when the agent detects either \texttt{MINI\_SWE\_AGENT\_\allowbreak FINAL\_OUTPUT} or \texttt{COMPLETE\_TASK\_AND\_SUBMIT\_\allowbreak FINAL\_OUTPUT}, it does not finalize the task immediately. Instead, it raises a \texttt{ContextRequested} exception, which triggers a structured context request. In the second stage, the agent validates the returned context format and proceeds with final submission only after successful verification.

\subsubsection{Format Validation Mechanism}

To enforce strict structure and machine-readability, the prompt is augmented with predefined templates that explicitly constrain the output format. Format validation is performed by a dedicated verification function, and a regular expression is used to extract the standardized \texttt{<PATCH\_CONTEXT>} block for downstream processing.

This design enforces strict structural constraints on contextual information, ensuring consistency, reproducibility, and compatibility with automated verification and downstream processing pipelines in agent-based software engineering workflows.

\subsubsection{Cross-Model Evaluation Results}

After establishing the context extraction framework within mini-SWE-agent, we evaluated its effectiveness across four distinct models: Claude Sonnet 4.5, GPT-5, Gemini 2.5 Pro, and DevStral2. 
Under this standardized extraction protocol, all four models demonstrated robust adherence to the framework, consistently producing accurate and machine-verifiable \texttt{<PATCH\_CONTEXT>} information. 
These results indicate that the context extraction mechanism effectively constrains model outputs, ensuring that the returned context is both structured and semantically relevant for downstream task submission.

During the evaluation, we observed that DevStral2 occasionally exhibited unexpected behaviors that deviated from the standardized output protocol. 
These anomalies are discussed in detail in~\cref{sec:appendix:devstral2}, where we provide the corresponding raw interaction traces and highlight the specific issues encountered.
\clearpage

\begin{tcolorbox}[breakable, colback=macaronblue!15, colframe=macaronblue!70, title={Context Control Templates in mini-SWE-Agent (context\_aware.yaml)}]
\small
\texttt{
context\_request\_template: | \\
 \\ \ <context\_request> \\
 \\ \ Before finalizing your submission, please provide the CODE CONTEXT you used to generate the patch. \\
}

\begin{tcolorbox}[breakable, colback=macaronblue!8, colframe=macaronblue!60, title={\textbf{Requirements}}, left=4pt, right=4pt, top=4pt, bottom=4pt]
\small
\texttt{
\ \ \ List ONLY the specific code sections from source files that you examined to create your patch.
}
\end{tcolorbox}

\begin{tcolorbox}[breakable, colback=macaronblue!8, colframe=macaronblue!60, title={\textbf{Format (strictly follow)}}, left=4pt, right=4pt, top=4pt, bottom=4pt]
\small
\texttt{
\ \ \ Each entry must include: \\
\ \ \ - File: <absolute\_file\_path> \\
\ \ \ - Lines: <start\_line>-<end\_line>
}
\end{tcolorbox}

\begin{tcolorbox}[breakable, colback=macaronblue!8, colframe=macaronblue!60, title={\textbf{Example}}, left=4pt, right=4pt, top=4pt, bottom=4pt]
\small
\texttt{
\ \ \ File: /testbed/src/core/handler.ext \\
\ \ \ Lines: 34-56 \\
\ \ \ \\
\ \ \ File: /testbed/lib/utils.ext \\
\ \ \ Lines: 128-145 \\
\ \ \ \\
\ \ \ File: /testbed/src/parser.ext \\
\ \ \ Lines: 67-89
}
\end{tcolorbox}

\begin{tcolorbox}[breakable, colback=macaronblue!8, colframe=macaronblue!60, title={\textbf{DO NOT Include}}, left=4pt, right=4pt, top=4pt, bottom=4pt]
\small
\texttt{
\ \ \ - Explanations, reasoning, or commentary \\
\ \ \ - Error messages or debugging output \\
\ \ \ - Test code or test results \\
\ \ \ - Documentation or comments \\
\ \ \ - Any non-code content
}
\end{tcolorbox}

\texttt{
\ \ \ Please format your response as: \\
\ \ \ \\
\ \ \ THOUGHT: Providing code context used for patch generation. \\
\ \ \ \\
\ \ \ <PATCH\_CONTEXT> \\
\ \ \ File: <absolute\_file\_path> \\
\ \ \ Lines: <start\_line>-<end\_line> \\
\ \ \ \\
\ \ \ File: <absolute\_file\_path> \\
\ \ \ Lines: <start\_line>-<end\_line> \\
\ \ \ </PATCH\_CONTEXT> \\
\ \ \ ```bash \\
\ \ \ echo COMPLETE\_TASK\_AND\_SUBMIT\_FINAL\_OUTPUT \&\& (git status >/dev/null 2>\&1 \&\& git add -A \&\& git diff --cached) || (cd */. 2>/dev/null \&\& git add -A \&\& git diff --cached) || (echo "Error: No git repository found") \\
\ \ \ ``` \\
\ \ \ \\
\ \ \ **CRITICAL:** Provide ONLY file paths and line ranges (no code snippets, no explanations) within <PATCH\_CONTEXT> tags. \\
\ \ \ </context\_request> \\
\ \ \ \\
context\_confirmation\_template: | \\
\ \ \ <context\_received> \\
\ \ \ Context received and recorded (\{\{context\_length\}\} characters). \\
\ \ \ Now proceeding with final submission. \\
\ \ \ </context\_received>
}
\end{tcolorbox}
% Space for filling experimental results and observations.

\begin{tcolorbox}[breakable, colback=macaronblue!15, colframe=macaronblue!70, title={Experiment Setting}]
\small
\texttt{
environment: \\
    timeout: 180 \\
    pull\_timeout: 3600 \\
    env: \\
        PAGER: cat \\
        MANPAGER: cat \\
        LESS: -R \\
        PIP\_PROGRESS\_BAR: 'off' \\
        TQDM\_DISABLE: '1' \\
    environment\_class: docker \\
model: \\
  model\_name: "anthropic/claude-sonnet-4-5-20250929" \\
  model\_kwargs: \\
    drop\_params: true \\
    temperature: 0.0
}
\end{tcolorbox}

\subsection{Cross-Agent Comparison for the Same Model}
\label{sec:appendix:a1.2}

\subsubsection{Context Extraction Protocol Extension}

To evaluate the impact of agent architecture and prompt design on context extraction, we extended the standardized context enforcement protocol, originally developed for mini-SWE-agent, to multiple other agents, including SWE-Agent, Openhands, and Prometheus. 
Each of these agents adopts a context extraction framework built upon the mini-SWE-agent as the foundational design, with agent-specific adaptations to accommodate differences in workflow and output formatting.

In all systems, an explicit pre-submission constraint was introduced, requiring the agent to output the precise code context prior to executing the final submission call. 
This ensures that all inspected source files and line ranges are consistently logged, enabling reproducibility, auditing, and automated verification of agent behavior.

Moreover, prompt templates and verification routines were adjusted for each agent to maintain a uniform, machine-readable \texttt{<PATCH\_CONTEXT>} structure. 
These adaptations facilitate a controlled comparison of context extraction effectiveness across heterogeneous agent frameworks, and provide a foundation for analyzing the influence of agent-specific design choices on model performance.

\subsubsection{SWE-Agent Implementation}

\begin{tcolorbox}[breakable, colback=macaronblue!15, colframe=macaronblue!70, title={Pre-Submission Context Enforcement Specification in SWE-Agent}]
\small
\textbf{Required Format:}
In the next response, the agent must:
\begin{enumerate}
    \item Output a \texttt{\textless PATCH\_CONTEXT\textgreater} block listing all source files and line ranges examined.
    \item Invoke the \texttt{submit} function.
\end{enumerate}

\begin{tcolorbox}[colback=macaronblue!8, colframe=macaronblue!60, title={\textbf{Format for \texttt{\textless PATCH\_CONTEXT\textgreater}}}, left=4pt, right=4pt, top=4pt, bottom=4pt, breakable]
\begin{verbatim}
<PATCH_CONTEXT>
File: /testbed/path/to/file1.ext
Lines: 10-50

File: /testbed/path/to/file2.ext
Lines: 100-150
</PATCH_CONTEXT>
\end{verbatim}
\end{tcolorbox}

\begin{tcolorbox}[colback=macaronblue!8, colframe=macaronblue!60, title={\textbf{Rules}}, left=4pt, right=4pt, top=4pt, bottom=4pt]
\begin{itemize}
    \item List only source files that were viewed or edited (exclude test files).
    \item Include the exact line ranges examined.
    \item Do not include code snippets, explanations, or commentary.
    \item This step is mandatory and cannot be skipped.
\end{itemize}
\end{tcolorbox}

\end{tcolorbox}

\begin{tcolorbox}[breakable, colback=macaronblue!15, colframe=macaronblue!70, title={config.yaml}]
\small
\texttt{
model: \\
    name: gpt-5 \\
    api\_base:  \\
    api\_key:  \\
    per\_instance\_cost\_limit: 0.0 \\
    total\_cost\_limit: 0.0 \\
    temperature: 1.0 \\
    top\_p: null \\
    retry: \\
      retries: 10 \\
      min\_wait: 5 \\
      max\_wait: 60 \\
    completion\_kwargs: \\
      timeout: 120 \\
  tools: \\
    execution\_timeout: 600 \\
    env\_variables: \\
      PAGER: cat \\
      MANPAGER: cat \\
      LESS: -R \\
      GIT\_PAGER: cat
}
\end{tcolorbox}

\subsubsection{Openhands Implementation}

\begin{tcolorbox}[colback=macaronblue!15, colframe=macaronblue!70, title={Context Extraction Protocol in Openhands (system\_prompt\_context\_aware.j2)}, breakable]

\textbf{Explore-context marking protocol.}
When exploring source code context (e.g., reading files or printing line ranges), the agent is required to emit a machine-parseable explore-context block before the corresponding command.

\begin{tcolorbox}[colback=macaronblue!8, colframe=macaronblue!60, title={\textbf{Explore Context Format}}, left=4pt, right=4pt, top=4pt, bottom=4pt]
\begin{lstlisting}[basicstyle=\ttfamily\small, frame=single]
<EXPLORE_CONTEXT>
File: /absolute/path/to/file.ext
Lines: <start>-<end>
</EXPLORE_CONTEXT>
\end{lstlisting}
\end{tcolorbox}

The explore-context block must contain one or more file-range entries with absolute paths and positive line indices. 
It is only included when the command prints source code content to stdout and must not be used for metadata-only commands (e.g., ls, grep, git status).

\medskip

% ===================== Nested Highlight Box =====================
\begin{tcolorbox}[
  colback=macaronblue!8,
  colframe=macaronblue!60,
  title={\textbf{Patch Context Submission Requirement}},
  left=4pt, right=4pt, top=4pt, bottom=4pt,
  breakable
]

Before finalizing the submission, the agent must provide the code context used for patch generation.

\textbf{Format (strictly enforced):}

\begin{lstlisting}[basicstyle=\ttfamily\small, frame=single]
THOUGHT: Providing code context used for patch generation.

<PATCH_CONTEXT>
File: <absolute_file_path>
Lines: <start_line>-<end_line>

File: <absolute_file_path>
Lines: <start_line>-<end_line>
</PATCH_CONTEXT>
\end{lstlisting}

Only absolute file paths and line ranges are permitted inside the \texttt{<PATCH\_CONTEXT>} block. 
Explanations, comments, debugging output, test code, or any non-code content are strictly disallowed.

\end{tcolorbox}
% ===================== End Nested Box =====================

\medskip

Unlike mini-SWE-agent and SWE-Agent, Openhands employs a different prompt format. 
Nevertheless, this variation in prompt design does not preclude the implementation of an equivalent context extraction mechanism. 
By adapting the extraction protocol to accommodate the agent-specific prompt structure, we ensure that the resulting \texttt{<PATCH\_CONTEXT>} outputs remain consistent, structured, and machine-verifiable, thereby preserving the integrity and comparability of the context extraction process across heterogeneous agents.

\end{tcolorbox}

\subsubsection{Agentless Enhancements}

\paragraph{Bug Fixes and Compatibility.}
Additionally, we implement several extensions and bug fixes for Agentless to ensure compatibility with the SWE-bench evaluation harness. We adapt to SWE-bench harness API changes by fixing import paths due to the test\_spec module restructuring. We also unify Docker environment configurations by removing hardcoded docker\_specs and aligning with build\_env\_images() to ensure consistent env\_image\_key generation. Additionally, we remove the deprecated pip option --no-use-pep517 which has been removed in pip $\geq$ 23.0.

\paragraph{Multi-Language Support.}
To support cross-language evaluation, we optimize the multi-language adaptation capability of Agentless, enabling better support for multi-language benchmarks.

\begin{tcolorbox}[breakable, colback=macaronblue!15, colframe=macaronblue!70, title={Improvement 1: Dynamic File Extension Adaptation}]
\small

\textbf{Strategy:} Automatically detect the repository's primary programming language by analyzing file extension frequency, then dynamically inject the most common extension into prompt examples.

\textbf{Detection Method:}
\begin{itemize}
    \item Traverse repository structure to count file extensions
    \item Select extension with highest occurrence (e.g., .py, .java, .js, .go)
    \item Replace hard-coded examples with detected extension
\end{itemize}

\vspace{0.3cm}

\begin{tcolorbox}[colback=macaronblue!8, colframe=macaronblue!60, title={\textbf{Before: Hard-coded Python Extensions}}, left=4pt, right=4pt, top=4pt, bottom=4pt]
\begin{verbatim}
For example:
```
file1.py
file2.py
```

### Examples:
```
full_path1/file1.py
line: 10
class: MyClass1

full_path2/file2.py
function: MyClass2.my_method
```
\end{verbatim}
\end{tcolorbox}

\begin{tcolorbox}[colback=macaronblue!8, colframe=macaronblue!60, title={\textbf{After: Repository-Adaptive Extensions}}, left=4pt, right=4pt, top=4pt, bottom=4pt]
\begin{verbatim}
For example:
```
file1.<detected_extension>
file2.<detected_extension>
```

### Examples:
```
full_path1/file1.<detected_extension>
line: 10
class: MyClass1

full_path2/file2.<detected_extension>
function: MyClass2.my_method
```
\end{verbatim}
\end{tcolorbox}

\textbf{Impact:} Eliminates language-specific bias in fault localization prompts, enabling cross-language applicability without manual configuration.
\end{tcolorbox}

\begin{tcolorbox}[breakable, colback=macaronblue!15, colframe=macaronblue!70, title={Improvement 2: Language-Agnostic DIFF Examples}]
\small

\textbf{Strategy:} Replace language-specific DIFF examples with universal "Hello World" modification patterns that adapt to the detected programming language.

\vspace{0.3cm}

\begin{tcolorbox}[colback=macaronblue!8, colframe=macaronblue!60, title={\textbf{Before: Python-Specific DIFF Example}}, left=4pt, right=4pt, top=4pt, bottom=4pt]
\begin{verbatim}
Here is an example:

```python
### mathweb/flask/app.py
<<<<<<< SEARCH
from flask import Flask
=======
import math
from flask import Flask
>>>>>>> REPLACE
```
\end{verbatim}
\end{tcolorbox}

\begin{tcolorbox}[colback=macaronblue!8, colframe=macaronblue!60, title={\textbf{After: Python Hello World Example}}, left=4pt, right=4pt, top=4pt, bottom=4pt]
\begin{verbatim}
Here is an example:

```python
### example/hello.py
<<<<<<< SEARCH
print("Hello World")
=======
print("Hello")
print("World")
>>>>>>> REPLACE
```
\end{verbatim}
\end{tcolorbox}

\begin{tcolorbox}[colback=macaronblue!8, colframe=macaronblue!60, title={\textbf{After: Java Hello World Example}}, left=4pt, right=4pt, top=4pt, bottom=4pt]
\begin{verbatim}
Here is an example:

```java
### example/Hello.java
<<<<<<< SEARCH
System.out.println("Hello World");
=======
System.out.println("Hello");
System.out.println("World");
>>>>>>> REPLACE
```
\end{verbatim}
\end{tcolorbox}

\begin{tcolorbox}[colback=macaronblue!8, colframe=macaronblue!60, title={\textbf{After: JavaScript Hello World Example}}, left=4pt, right=4pt, top=4pt, bottom=4pt]
\begin{verbatim}
Here is an example:

```javascript
### example/hello.js
<<<<<<< SEARCH
console.log("Hello World");
=======
console.log("Hello");
console.log("World");
>>>>>>> REPLACE
```
\end{verbatim}
\end{tcolorbox}

\begin{tcolorbox}[colback=macaronblue!8, colframe=macaronblue!60, title={\textbf{After: Go Hello World Example}}, left=4pt, right=4pt, top=4pt, bottom=4pt]
\begin{verbatim}
Here is an example:

```go
### example/hello.go
<<<<<<< SEARCH
fmt.Println("Hello World")
=======
fmt.Println("Hello")
fmt.Println("World")
>>>>>>> REPLACE
```
\end{verbatim}
\end{tcolorbox}

\textbf{Adaptation Logic:}
\begin{itemize}
    \item Detect repository language from file extensions
    \item Select corresponding Hello World DIFF template
    \item Inject into repair prompts at runtime
\end{itemize}

\textbf{Impact:} Provides language-appropriate syntax examples without domain-specific dependencies (e.g., Flask), improving LLM understanding across diverse codebases.
\end{tcolorbox}

\subsubsection{Framework Generalization Results}

After applying the standardized context extraction framework to all five agents—mini-SWE, SWE-Agent, Openhands, agentless, and Prometheus—we observed that each system was able to accurately produce the corresponding \texttt{<PATCH\_CONTEXT>} outputs. 
This demonstrates that the framework is robust and generalizable across heterogeneous agent architectures, effectively constraining model outputs to a structured, machine-verifiable format.

The consistent extraction of patch context across agents facilitates reproducibility and auditing, and provides a reliable basis for downstream evaluation of model performance in automated software engineering tasks. 
Moreover, these results indicate that the underlying principles of context enforcement can be adapted to accommodate differences in prompt design, agent workflow, and system architecture without compromising the fidelity or completeness of the extracted context. 
Overall, the successful deployment of the framework across multiple agents underscores its utility as a foundational tool for systematic evaluation and comparison of agent-based patch generation systems.

\section{Gold Context Definition and Scope}
\label{sec:appendix:gold_context_definition}

\parabf{Gold Context as a Compact Sufficient Reference.}
In \benchmark, we define the \emph{gold context} as a human-annotated set of code regions that is \emph{verified to be sufficient} for resolving the issue, while being \emph{as compact as possible} under our annotation guideline (the minimal context principle).
We do not claim global minimality: enforcing a globally minimal sufficient set is generally intractable in repository-scale codebases due to the combinatorial space of dependency chains and code spans.
Instead, our refinement stage removes clearly redundant regions while preserving sufficiency under an executable feasibility check (\cref{sec:appendix:detail_context_verification}).
Accordingly, precision/F1 should be interpreted with respect to this compact-and-verified reference, rather than as a strict penalty for retrieving any additional context that may also be helpful.

\clearpage
\section{Block-Level (AST) Alignment Details}
\label{sec:appendix:block_nodes}

\parabf{Block-Level (AST) Alignment.}
At the block level, we do not treat arbitrary AST nodes as evaluation units.
Instead, we standardize blocks as \emph{definition-level symbols} (e.g., function/method/class/interface/trait definitions) to ensure cross-language consistency.
Concretely, we parse each repository file with \texttt{tree-sitter} and extract a language-specific set of node types that correspond to top-level or member definitions (\autoref{tab:block_node_types}).
Each extracted node is mapped to a canonical block span by its file path and \texttt{(start\_line, end\_line)} range (converted from \texttt{tree-sitter} byte offsets).
This design ensures that “block” refers to comparable semantic units across languages (definitions that encapsulate reusable logic or APIs), rather than low-level syntax nodes whose granularity varies substantially across parsers.

\begin{table}[htbp]
\centering
\small
\setlength{\tabcolsep}{6pt}
\renewcommand{\arraystretch}{1.05}
\caption{\textbf{Tree-sitter node types used for block-level (definition-level) evaluation across languages.}
Blocks are standardized as definition-level symbols to improve cross-language comparability.}
\begin{tabular}{p{0.18\linewidth}p{0.76\linewidth}}
\toprule
\textbf{Language} & \textbf{Node types (block units)} \\
\midrule
Python & \texttt{function\_definition}, \texttt{class\_definition}, \texttt{async\_function\_definition} \\
JavaScript & \texttt{function\_declaration}, \texttt{class\_declaration}, \texttt{method\_definition}, \texttt{arrow\_function} \\
TypeScript & \texttt{function\_declaration}, \texttt{class\_declaration}, \texttt{method\_definition}, \texttt{interface\_declaration} \\
TSX & \texttt{function\_declaration}, \texttt{class\_declaration}, \texttt{method\_definition}, \texttt{interface\_declaration} \\
Java & \texttt{method\_declaration}, \texttt{class\_declaration}, \texttt{interface\_declaration}, \texttt{constructor\_declaration} \\
Go & \texttt{function\_declaration}, \texttt{method\_declaration}, \texttt{type\_declaration} \\
Rust & \texttt{function\_item}, \texttt{impl\_item}, \texttt{struct\_item}, \texttt{trait\_item} \\
C & \texttt{function\_definition}, \texttt{struct\_specifier} \\
C++ & \texttt{function\_definition}, \texttt{class\_specifier}, \texttt{struct\_specifier} \\
C\# & \texttt{method\_declaration}, \texttt{class\_declaration}, \texttt{interface\_declaration} \\
PHP & \texttt{function\_definition}, \texttt{method\_declaration}, \texttt{class\_declaration} \\
Ruby & \texttt{method}, \texttt{class}, \texttt{module} \\
Swift & \texttt{function\_declaration}, \texttt{class\_declaration}, \texttt{protocol\_declaration} \\
Kotlin & \texttt{function\_declaration}, \texttt{class\_declaration} \\
Scala & \texttt{function\_definition}, \texttt{class\_definition}, \texttt{trait\_definition} \\
\bottomrule
\end{tabular}
\label{tab:block_node_types}
\end{table}

\clearpage
\section{Details on Annotation}
\label{sec:appendix:a2}

To obtain precise problem-specific contextual annotations, we developed a dedicated annotation frontend that supports multiple annotators working in parallel with randomized task assignment. 
The interface provides lightweight IDE-like functionalities, including navigation via function and variable references, as well as object-oriented class definition lookup, enabling annotators to efficiently explore repository-level codebases. 

\begin{figure}[H]
    \centering
    \begin{minipage}{0.58\textwidth}
        \centering
        \includegraphics[width=\textwidth]{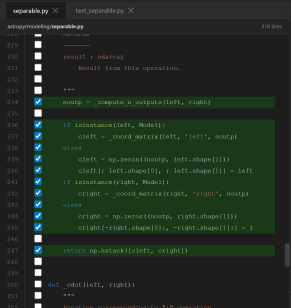}
        \caption*{(a) Code annotation view.}
    \end{minipage}
    \hfill
    \begin{minipage}{0.38\textwidth}
        \centering
        \includegraphics[width=\textwidth]{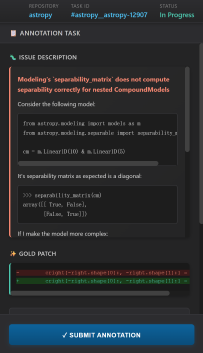}
        \caption*{(b) Sidebar with issue description and statistics.}
    \end{minipage}
    \caption{Annotation frontend UI. Annotators select minimal problem-specific code context while monitoring task-level metadata in the side panel.}
    \label{fig:annotation_ui}
\end{figure}

\autoref{fig:annotation_ui} presents the annotation frontend UI.
The code annotation view (left) allows annotators to select minimal problem-specific code spans with lightweight IDE-style navigation.
The side panel (right) displays structured task metadata, including issue descriptions, reference patches, and real-time annotation statistics, enabling annotators to maintain awareness of the overall task scope.

For a given annotation task, annotators are blinded to each other's annotations to prevent information leakage and anchoring bias.

We refer readers to the main paper for the full annotation and verification pipeline.This multi-round blind annotation protocol ensures both correctness and robustness of the resulting gold context dataset. 

Our annotator team consists of highly trained software engineering experts with advanced research experience in large-scale codebases.

All annotators underwent a unified training process before participating in the annotation tasks. 
During training, we enforced a standardized annotation guideline, requiring annotators to provide precise and minimal contextual information. 
Specifically, annotators were instructed to follow a \emph{minimal context principle}, selecting only the code segments that were strictly necessary for resolving the issue, and avoiding redundant or irrelevant code regions.

The annotation time per issue ranged from approximately 20 minutes to 1.5 hours, depending on the complexity of the repository and the issue scope. 
On average, producing a high-quality annotation for a single issue in a repository-scale project requires approximately 40 minutes of effort by an expert annotator.

For repositories with exceptionally large or complex codebases, we imposed an upper bound on annotation time to control annotation cost and prevent diminishing returns from excessively long inspection sessions.

We periodically audited annotations to ensure guideline compliance and consistency across annotators.

\clearpage
\section{Details on Context Verification}
\label{sec:appendix:detail_context_verification}

This subsection provides additional details on the context verification procedure used in our analysis.
The goal of this process is to determine whether an annotated context contains sufficient information
to resolve its corresponding issue, rather than to evaluate or compare the repair capabilities of
specific language models.

\paragraph{Verification Setup.}
For each task, we condition a strong state-of-the-art language model (e.g., GPT-5) solely on the
annotated context, without access to any external files, repositories, or retrieval mechanisms.
The model is prompted to generate multiple independent candidate patches (five attempts in our
experiments), each following the same prompt template.
All generated patches are evaluated using the official test suite associated with the task,
including both fail-to-pass and pass-to-pass test cases.

\paragraph{Sufficiency Criterion.}
An annotated context is considered sufficient if at least one of the generated patches passes the
entire test suite.
This criterion is intentionally defined in terms of existence rather than consistency:
we do not require the model to reliably solve the task across all attempts.
Instead, we only check whether a valid solution is possible given the provided context.
This aligns with our objective of verifying context sufficiency, rather than measuring average
model performance.

\paragraph{Rationale.}
Using a strong language model in this setting serves as a conservative lower-bound feasibility check.
If even a capable agent fails to produce a correct fix when restricted to the annotated context,
it is unlikely that the context alone contains all information required to resolve the issue.
Conversely, a successful patch indicates that the annotated context is, in principle, sufficient
to support a correct solution.

\paragraph{Sampling and Stochasticity.}
We employ multiple independent generations to mitigate the effects of stochastic decoding and
occasional generation failures.
This design reduces false negatives caused by randomness, while avoiding over-reliance on a
single generation outcome.
Importantly, correctness is determined exclusively by the test suite, independent of the model's
internal confidence or reasoning traces.

Overall, this verification procedure should be interpreted as a feasibility-oriented validation
of annotated contexts, rather than a benchmark of language model repair performance.

\clearpage
\section{Details on Evaluation Metrics}
\label{sec:appendix:metrics}

% \subsection{Evaluation Metrics}
We evaluate coding agents by quantifying how effectively they retrieve relevant context compared to the human-verified gold contexts.
Formally, given an issue task $T$, let $C^{G}$ denote the gold context, $C^{A}$ denote the final context retrieved by the agent, and $C^{A}_{i}$ represent the intermediate context snapshot collected at the $i$-th checkpoint during the agent’s execution trajectory.
All $C^{G}$, $C^{A}$, and $C^{A}_{i}$ are represented as sets of code elements at a specified granularity (\textit{e.g.}, file, AST block, or line).

%notation
% For a given issue instance, let $G$ denote the \emph{gold context} and $A$ denote the \emph{agent-retrieved context}. Both $G$ and $A$ are represented as sets of code elements under a given granularity (file, block, or line).

\parabf{Context Recall.}
This metric quantifies the proportion of the gold context required for issue resolution that is successfully retrieved by the agent, formulated as:
\begin{equation}
\mathrm{Recall}(C^{A}, C^{G}) = \frac{|C^{A} \cap C^{G}|}{|C^{G}|}.
\end{equation}
A higher recall indicates that the coding agent is more effective at covering the essential context, whereas a lower score suggests that critical information is missing from its retrieved context.

\parabf{Context Precision.}
This metric quantifies the proportion of the agent’s retrieved context that overlaps with the gold context, defined as:
\begin{equation}
\mathrm{Precision}(C^{A}, C^{G}) = \frac{|C^{A} \cap C^{G}|}{|C^{A}|}.
\end{equation}
A higher precision indicates that the coding agent retrieves relevant and concise context, whereas a lower score suggests over-retrieval with redundant or irrelevant information that may introduce noise and hinder reasoning.

\parabf{Context F1.}
Context recall and precision capture two complementary aspects of an agent’s context retrieval behavior.
High recall with low precision indicates over-retrieval, where the agent includes excessive irrelevant context, while high precision with low recall indicates under-retrieval, where essential information is missed.
To provide a balanced evaluation, we compute the F1 score as the harmonic mean of precision and recall, defined as:
\begin{equation}
F_1 = \frac{2 \times \mathrm{Precision} \times \mathrm{Recall}}{\mathrm{Precision} + \mathrm{Recall}}.
\end{equation}

% \parabf{Context Retrieval Efficiency and Redundancy.}
% We also compute two process-oriented metrics to assess the intermediate context retrieval behavior of agents:
% \begin{equation}
%     \mathrm{Efficiency} = \frac1T \mathrm{Recall}(\cup_{i=1}^T C^A_i, C^G), \\ \mathrm{Redundancy} = \frac{|\cap_{i=1}^T C^A_i|}{}
% \end{equation}
% the cumulative gold context coverage 
% This metric measures how efficiently an agent retrieves critical contexts for issue resolution over time.
% Specifically, we plot the cumulative coverage of gold context during agent execution against the number of execution steps and compute the normalized area under the curve as:
% \begin{equation}
%     \mathrm{AUC}_{\mathrm{PCC}} = \frac1T \sum_{i=1}^T \mathrm{Recall}(C^{A}_i, C^{G}).
% \end{equation}
% A higher value indicates that the agent can retrieve necessary contexts earlier during the resolution process.

\parabf{Context Retrieval Efficiency.}
Beyond the final retrieved context $C^{A}$, we evaluate retrieval as a \emph{process} along the trajectory.
For each run, we log every file-observation action and construct a sequence of intermediate context snapshots $\{C^{A}_{t}\}_{t=1}^{T}$, where $C^{A}_{t}$ is the set of code elements (at a chosen granularity) inspected at step $t$.
We define the \emph{cumulative explored context} up to step $t$ as
\begin{equation}
A^{(t)} ~=~ \bigcup_{i=1}^{t} C^{A}_{i}.
\end{equation}
We then track the \emph{cumulative gold coverage} curve $\mathrm{Coverage}(A^{(t)}, C^{G})$ over $t=1,\dots,T$ and summarize how \emph{early} an agent reaches high gold coverage using the normalized area under this curve:
\begin{equation}
\mathrm{AUC\mbox{-}Cov} ~=~ \frac{1}{T} \sum_{t=1}^{T} \mathrm{Recall}(A^{(t)}, C^{G}).
\end{equation}
A higher AUC-Cov indicates that the agent retrieves critical gold evidence earlier during execution (i.e., reaches high coverage with fewer observation steps), whereas a lower value suggests that the agent either misses gold evidence or only discovers it late after many iterations.

\parabf{Context Retrieval Redundancy.}
While AUC-Cov captures how quickly coverage increases, it does not measure whether an agent repeatedly inspects the \emph{same} regions.
We therefore quantify redundancy by measuring how much of each newly retrieved snapshot overlaps with what has already been observed.
Formally, for $t \ge 2$, we define the \emph{per-step redundancy ratio} as the fraction of elements in $C^{A}_{t}$ that have appeared in previous steps:
\begin{equation}
\mathrm{Redun}_{t} ~=~ \frac{\left| C^{A}_{t} \cap \left(\bigcup_{i=1}^{t-1} C^{A}_{i}\right) \right|}{\left| C^{A}_{t} \right|}.
\end{equation}
We report the overall redundancy as the average across steps:
\begin{equation}
\mathrm{Redun} ~=~ \frac{1}{T-1} \sum_{t=2}^{T} \mathrm{Redun}_{t}.
\end{equation}
A higher redundancy indicates that an agent spends more retrieval budget revisiting previously seen context (i.e., looping or re-reading), whereas a lower redundancy indicates more \emph{novel} context acquisition per step.

% \parabf{Relationship to Final Context.}
% In addition to process metrics computed from $\{C^{A}_{t}\}$, we also evaluate the agent's final aggregated patch context $\tilde{A}$ (explicitly declared before submission) by computing $\mathrm{Recall}(\tilde{A}, C^{G})$, $\mathrm{Precision}(\tilde{A}, C^{G})$, and $F_{1}$ under the same granularity.
% This enables us to compare what the agent \emph{eventually observes} (via $A^{(t)}$) against what it \emph{ultimately retains for patch generation} (via $\tilde{A}$).

\parabf{Evidence Drop (Retrieval $\neq$ Use).}
High explored coverage does not necessarily translate to effective patching, because agents may \emph{observe} gold-relevant regions during exploration but fail to \emph{retain} them in the final aggregated patch context $\tilde{A}$.
To quantify this gap, we define the \emph{explored-gold set} as the gold elements that have been observed at least once during the trajectory:
\begin{equation}
G_{\mathrm{seen}} ~=~ \left(\bigcup_{t=1}^{T} C^{A}_{t}\right) \cap C^{G}.
\end{equation}
We then measure how much of this observed gold evidence is kept in the final patch context by
\begin{equation}
\mathrm{Keep} ~=~ \frac{\left| C^{A} \cap C^{G} \right|}{\left| G_{\mathrm{seen}} \right|},
\end{equation}
and define \emph{evidence drop} as its complement:
\begin{equation}
\mathrm{Drop} ~=~ 1 - \mathrm{Keep}
~=~ 1 - \frac{\left| C^{A} \cap C^{G} \right|}{\left| \left(\bigcup_{t=1}^{T} C^{A}_{t}\right) \cap C^{G} \right|}.
\end{equation}
A lower Drop (higher Keep) indicates that the agent successfully consolidates and preserves gold-relevant evidence it has already discovered, whereas a higher Drop indicates that gold evidence is frequently \emph{observed but discarded} during consolidation, highlighting failures in localization and evidence selection rather than pure recall-oriented retrieval.

\parabf{Gold Context Robustness.}
Our gold contexts are human-annotated based on a gold patch. However, in practice, a single issue may admit multiple semantically equivalent but syntactically different patches, raising the question of whether the collected gold context is reliable under solution multiplicity.
To answer this question, we analyze how gold contexts vary across multiple distinct test-passing patches for the same issue. We conduct a case study on 82 instances from \benchmark. For each instance, we collect two or three semantically equivalent patches and derive a patch-conditioned gold context for each patch using the same patch-driven context tracing procedure.
Let $\mathcal{G}=\{G^{(1)},\dots,G^{(K)}\}$ denote the set of gold contexts for a given instance. We quantify consistency using the Jaccard similarity (\autoref{jaccard}) between each pair of contexts. For each instance, we report the average similarity across all context pairs.
\begin{equation}
\label{jaccard}
\mathrm{Jaccard}(G^{(i)},G^{(j)}) ~=~ \frac{|G^{(i)} \cap G^{(j)}|}{|G^{(i)} \cup G^{(j)}|}.
\end{equation}

Note that all reported metrics are averaged over evaluated tasks.
Each metric can be computed at three granularities: file, block, and line levels.
By evaluating context retrieval across multiple metrics and granularities, \benchmark provides a more fine-grained understanding of agent behavior, beyond conventional end-to-end evaluation metrics that focus solely on final task success rates.

% Context retrieval can be evaluated at different levels of granularity. To provide a fine-grained and diagnostic assessment, we instantiate the above metrics at three complementary levels: file-level, block-level, and line-
% By evaluating context retrieval at multiple granularities, \benchmark provides a diagnostic assessment of agent behavior that is not captured by outcome-based benchmarks focusing solely on final task success.

% \subsection{Evaluation Settings.}
% Unless otherwise specified, each agent is run under its standard benchmark configuration. We sample a single solution per instance (\texttt{pass@1}) in the main results, and report additional \texttt{pass@k} analyses in supplementary experiments. We use a fixed decoding configuration (temperature and other generation parameters) across agents when applicable, and distinguish reasoning and non-reasoning model backbones where relevant.

\clearpage
\section{Details on Empirical Study}
\label{sec:appendix:detail_empirical_study}

This subsection provides implementation-level details of our empirical analysis.
To analyze agent behavior beyond final patch correctness, we instrument the evaluation pipeline to record intermediate artifacts, including tool-call logs, retrieved file and element contexts, and step-by-step edit trajectories.

Based on these records, we perform manual inspection to trace how each agent localizes relevant files, retrieves semantic information, and arrives at its final fix.
All annotation and analysis are conducted with the support of the annotation frontend UI introduced earlier, which enables structured inspection and labeling of the recorded artifacts.
The following case studies illustrate this analysis procedure for different agents, highlighting how the recorded artifacts are used to diagnose failure points at various stages of the debugging process.

\subsection{Case 1: Prometheus Agent – Incomplete Class Semantics Retrieval}
\label{case:Prometheus}

\textbf{Task}: Fix issue \texttt{psf\_\_requests-1921} where setting session headers to \texttt{None} sends literal "None" instead of omitting the header.

\textbf{Diagnostic Findings}:
\begin{enumerate}
    \item \textbf{Partial Class Retrieval}: Agent logs show successful retrieval of \texttt{CaseInsensitiveDict}'s \texttt{\_\_setitem\_\_} and \texttt{\_\_getitem\_\_} methods (lines 71-78), but \emph{no retrieval} of its \texttt{\_\_init\_\_} or \texttt{update} methods.
    
    \item \textbf{Semantic Mismatch}: The generated patch passed a generator expression to \texttt{CaseInsensitiveDict} (line 54 of \texttt{requests/sessions.py}):
    \begin{lstlisting}[language=Python]
merged_setting = dict_class((k, v) for k, v in ... if v is not None)
    \end{lstlisting}
    Analysis of the missing \texttt{\_\_init\_\_} reveals it requires a mapping with \texttt{.items()}, not a generator.
    
    \item \textbf{Impact}: 26 test failures due to dropped headers, broken authentication, and cookie persistence failures.
\end{enumerate}

\textbf{Gold Patch Comparison}: The correct fix adds a single filtering line \emph{after} merging, preserving proper initialization:
\begin{lstlisting}[language=Python]
merged_setting = dict((k, v) for (k, v) in merged_setting.items() if v is not None)
\end{lstlisting}

\textbf{Root Cause}: \textbf{Element-level slicing insufficiency}—the agent retrieved operational methods but missed constructor semantics, leading to API contract violation.

\subsection{Case 2: Agentless – Failure at File Localization Stage}
\label{case:Agentless}

\textbf{Task}: Fix for \texttt{django\_\_django-11630}.

\textbf{Diagnostic Findings}:
\begin{enumerate}
    \item \textbf{Stage 1 Failure (File Localization)}: The agent 
    retrieved 10 files, including 5 from \texttt{django/db/models/*} 
    (e.g., \texttt{base.py}, \texttt{options.py}, \texttt{fields/}) 
    and 5 framework configuration files. Critically, 
    \texttt{django/core/checks/model\_checks.py}—the source of error 
    \texttt{E028} based on gold patch analysis—was \textbf{never 
    retrieved}.
    
    \item \textbf{Cascading Failure}: With incorrect files localized 
    in Stage 1, subsequent element localization (Stage 2) focused on 
    \texttt{ModelBase}, \texttt{Options} classes, and edit 
    localization (Stage 3) modified \texttt{options.py}'s 
    \texttt{db\_table} calculation. The gold patch, by contrast, 
    modifies logic in \texttt{model\_checks.py} to conditionally 
    emit warnings when database routers are configured.
    
    \item \textbf{Information-Architecture Gap}: The issue mentions 
    ``\texttt{db\_table} collision'' (symptom-level) but not the 
    validation framework (implementation-level). File tree shows 
    \texttt{core/checks/} exists but lacks semantic annotations. 
    Without backward-tracing from error code \texttt{E028} to its 
    source, the agent inferred the problem resided in model 
    definition layers.
\end{enumerate}

\textbf{Root Cause}: \textbf{File localization failure propagated 
through all stages}. Operating on surface-level keywords from the 
issue (``models'', ``db\_table''), the agent lacked mechanisms to:
(i) trace error codes to implementation sources, 
(ii) recognize Django's architectural separation of validation from 
model logic. This initial misstep rendered all downstream 
localization (element and edit) ineffective, as they operated on 
an incorrect file set.

\subsection{Case 3: OpenHands Agent – Cross-Context Exploration Gap}
\label{case:OpenHands}

\textbf{Task}: Fix for \texttt{django\_\_django-11138} regarding ignored \texttt{TIME\_ZONE} values in multi-backend configurations.

\textbf{Diagnostic Findings}:
\begin{enumerate}
    \item \textbf{Keyword-Centric Search Bias}: The agent relied heavily on \texttt{grep} for string matching. Since this specific SQL function is unique to MySQL, the tool’s output anchored the agent’s search space to the MySQL implementation, effectively shielding the relevant SQLite and Oracle modules from the localization process.
    
    \item \textbf{Structural Observation-Action Gap}: While \texttt{ls -R} tool calls explicitly listed the \texttt{sqlite3/} and \texttt{oracle/} directories, the agent's heuristic failed to trigger follow-up audit calls (e.g., \texttt{view} or targeted \texttt{grep}) for these parallel modules. This represents a failure to translate structural awareness into diagnostic action.
    
    \item \textbf{Cross-Backend Reproduction Paradox}: Faced with environment constraints for MySQL, the agent used \texttt{bash} tool calls to successfully run a reproduction script in a SQLite environment. Although the tool-verified evidence confirmed the bug's presence in SQLite, the localization logic failed to propagate this feedback, resulting in a ``reproduce in SQLite, fix in MySQL'' logical disconnect.
    
    \item \textbf{Canonical Pattern Neglect}: Despite retrieving the base operations class via \texttt{view}, the agent bypassed the framework's canonical \texttt{connection.timezone\_name} property. Instead, it utilized tool calls to inject non-standard dictionary-access logic (\texttt{settings\_dict.get('TIME\_ZONE')}), failing to align with Django's architectural standards.
\end{enumerate}

\textbf{Root Cause}: \textbf{Search-Induced Context Tunneling}. The agent's localization strategy is highly sensitive to initial tool-call results but lacks \textbf{semantic generalization}. High-relevance hits in one module (MySQL) acted as a distractor, suppressing the horizontal exploration of parallel modules (SQLite/Oracle). This reliance on keyword-driven slicing leads to systemic fix incompleteness in modular architectures where logic is distributed across similar but syntactically distinct files.

\clearpage
\section{Cross-Language Radar Distributions for File/Block/Line Metrics}
\label{sec:appendix:radar_lang}

% Block = Symbol, Line = Span (rename here if your paper defines differently)
\begin{figure*}[htbp]
    \centering
    \setlength{\tabcolsep}{2pt}
    \renewcommand{\arraystretch}{1.0}

    \begin{tabular}{ccc}
        \subcaptionbox{\textbf{File} Precision}{\includegraphics[width=0.30\textwidth]{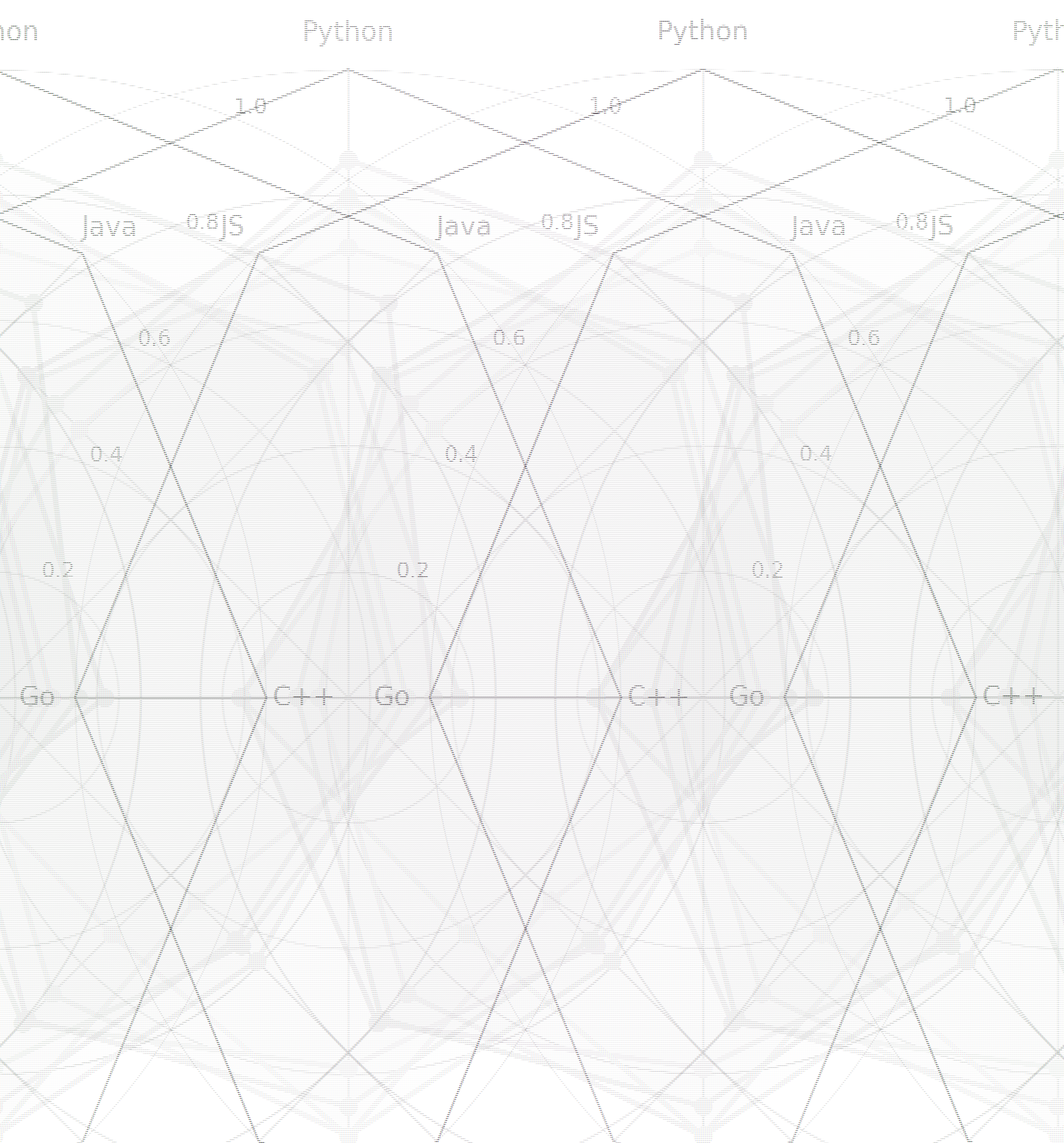}} &
        \subcaptionbox{\textbf{Block} Precision}{\includegraphics[width=0.30\textwidth]{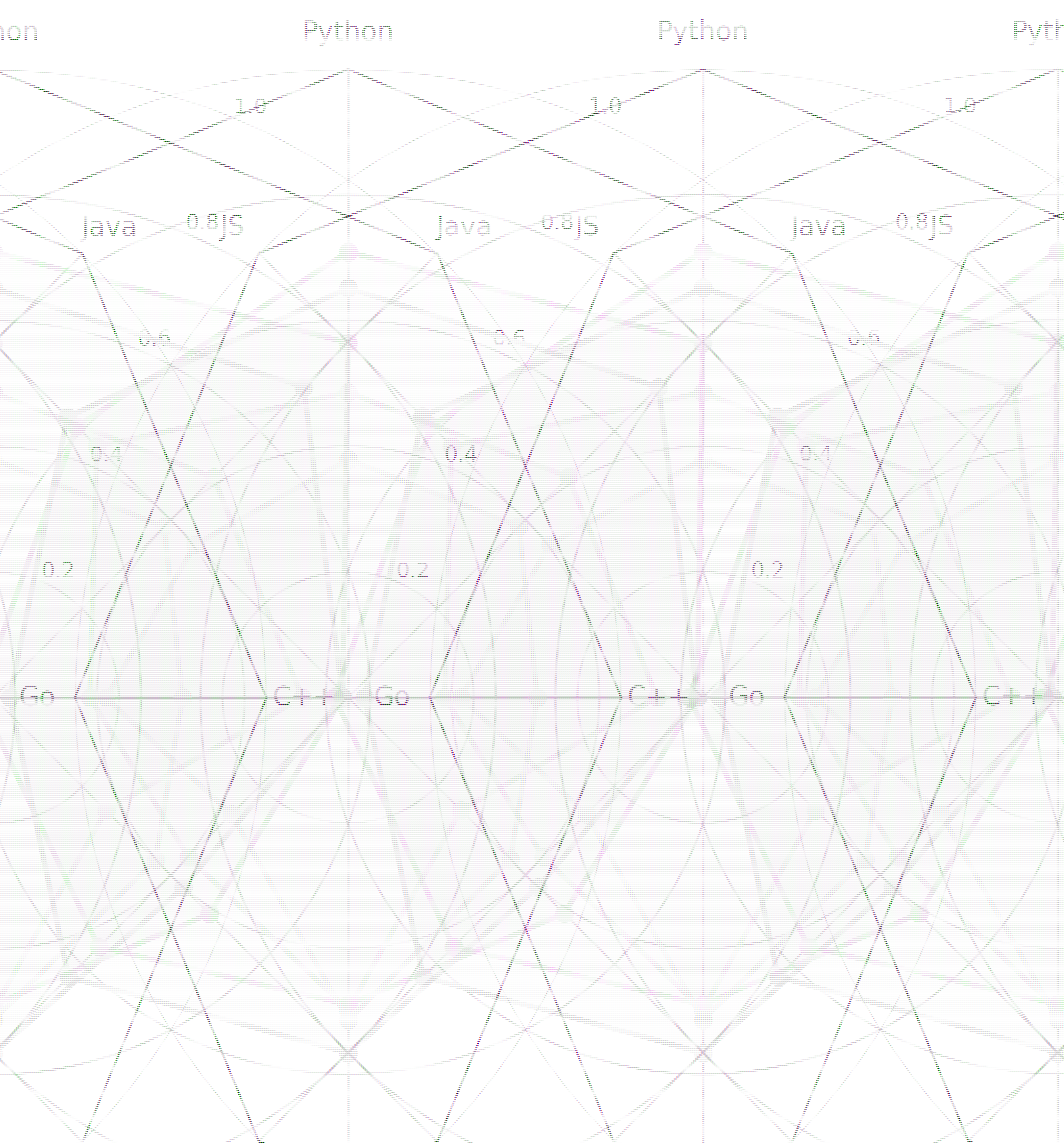}} &
        \subcaptionbox{\textbf{Line} Precision}{\includegraphics[width=0.30\textwidth]{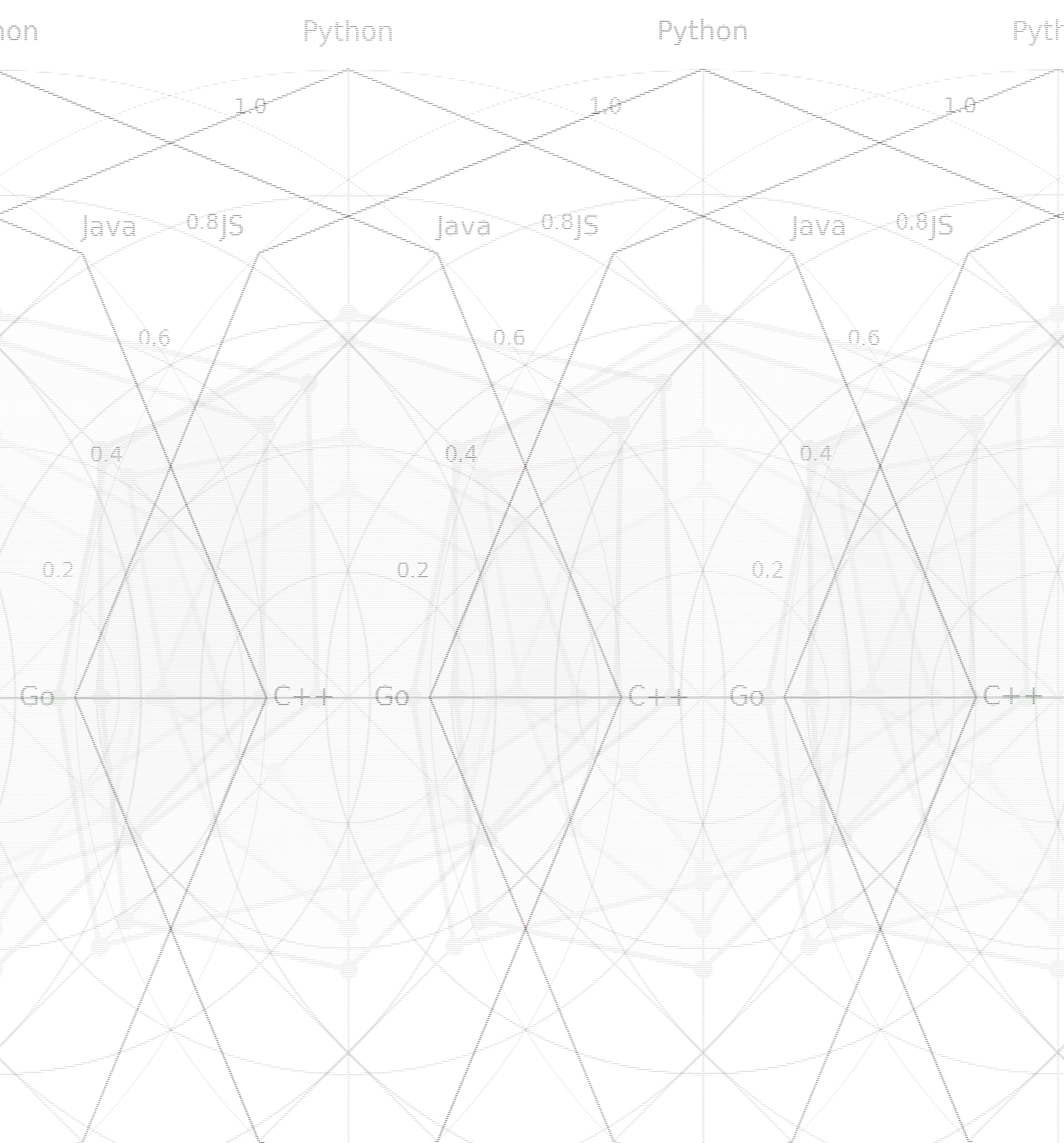}} \\

        \subcaptionbox{\textbf{File} Recall}{\includegraphics[width=0.30\textwidth]{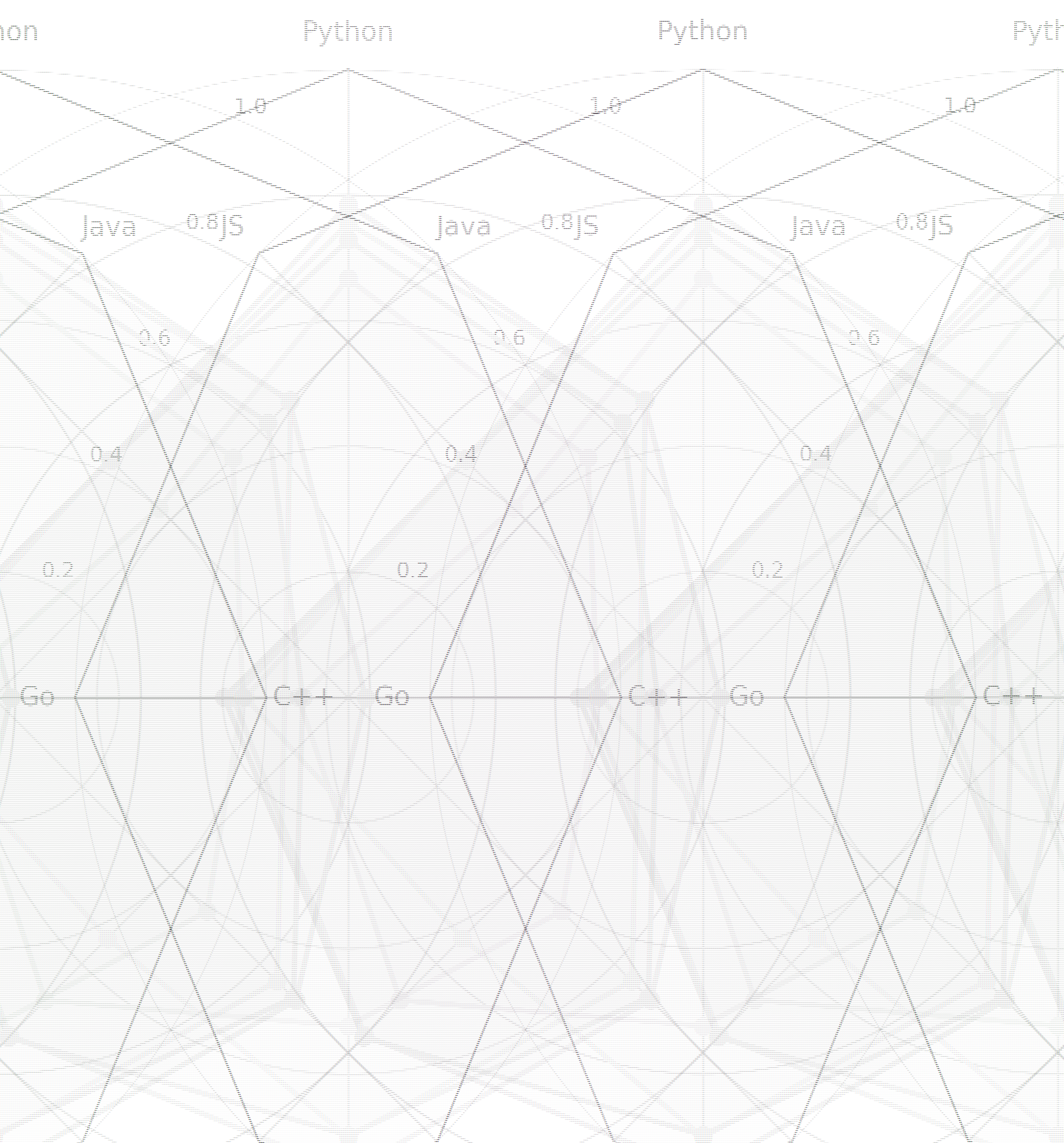}} &
        \subcaptionbox{\textbf{Block} Recall}{\includegraphics[width=0.30\textwidth]{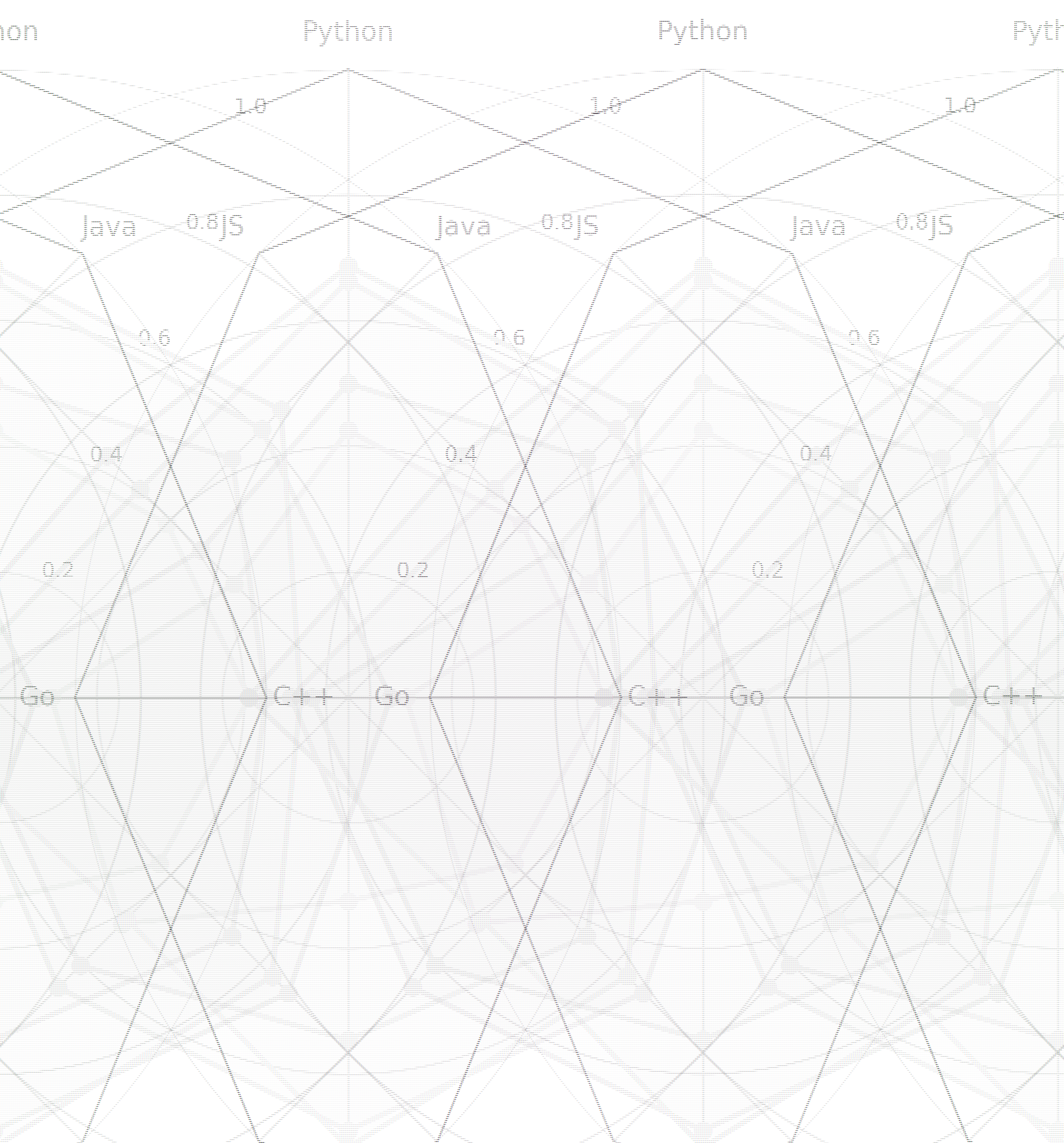}} &
        \subcaptionbox{\textbf{Line} Recall}{\includegraphics[width=0.30\textwidth]{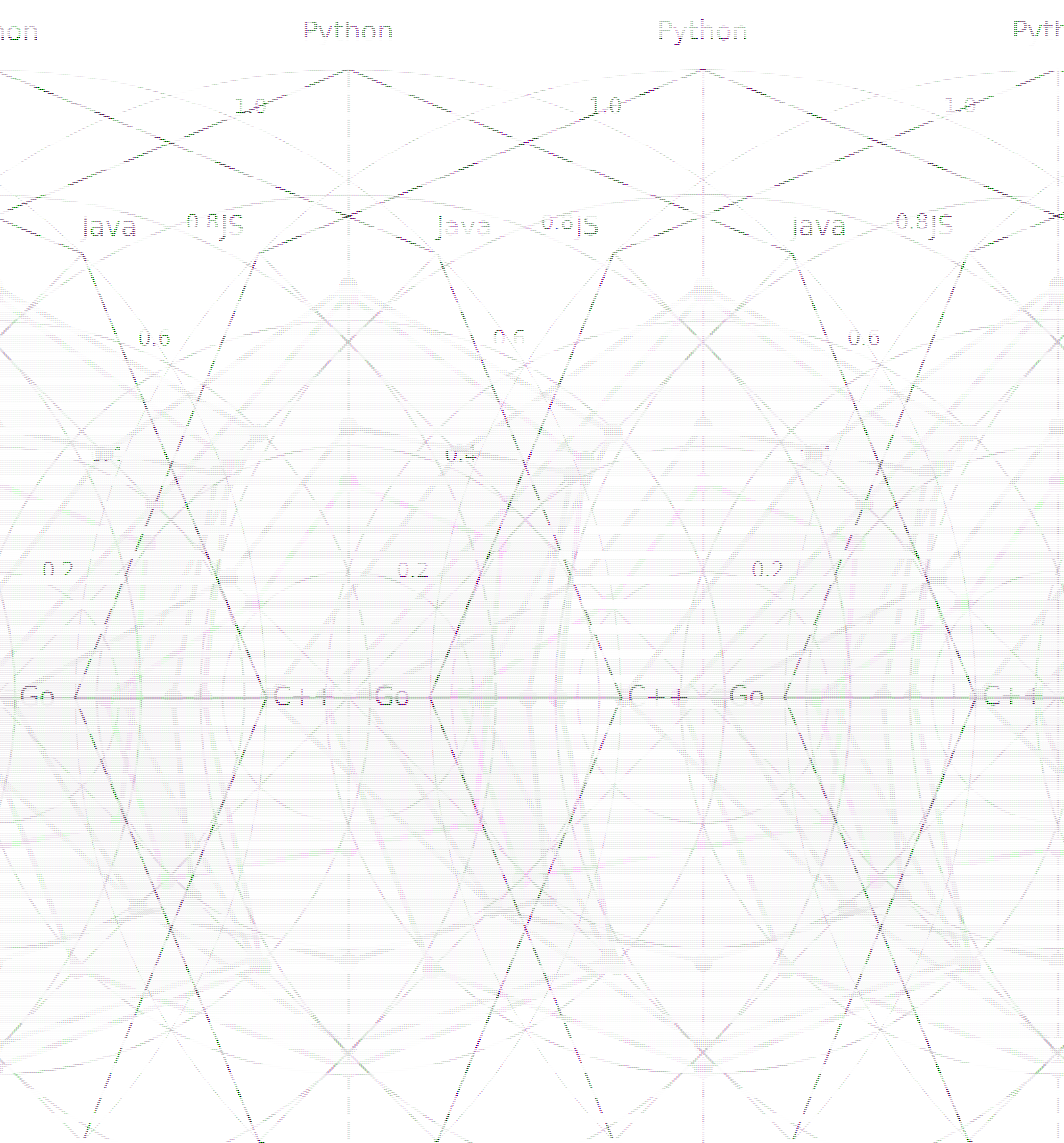}} \\

        \subcaptionbox{\textbf{File} F1}{\includegraphics[width=0.30\textwidth]{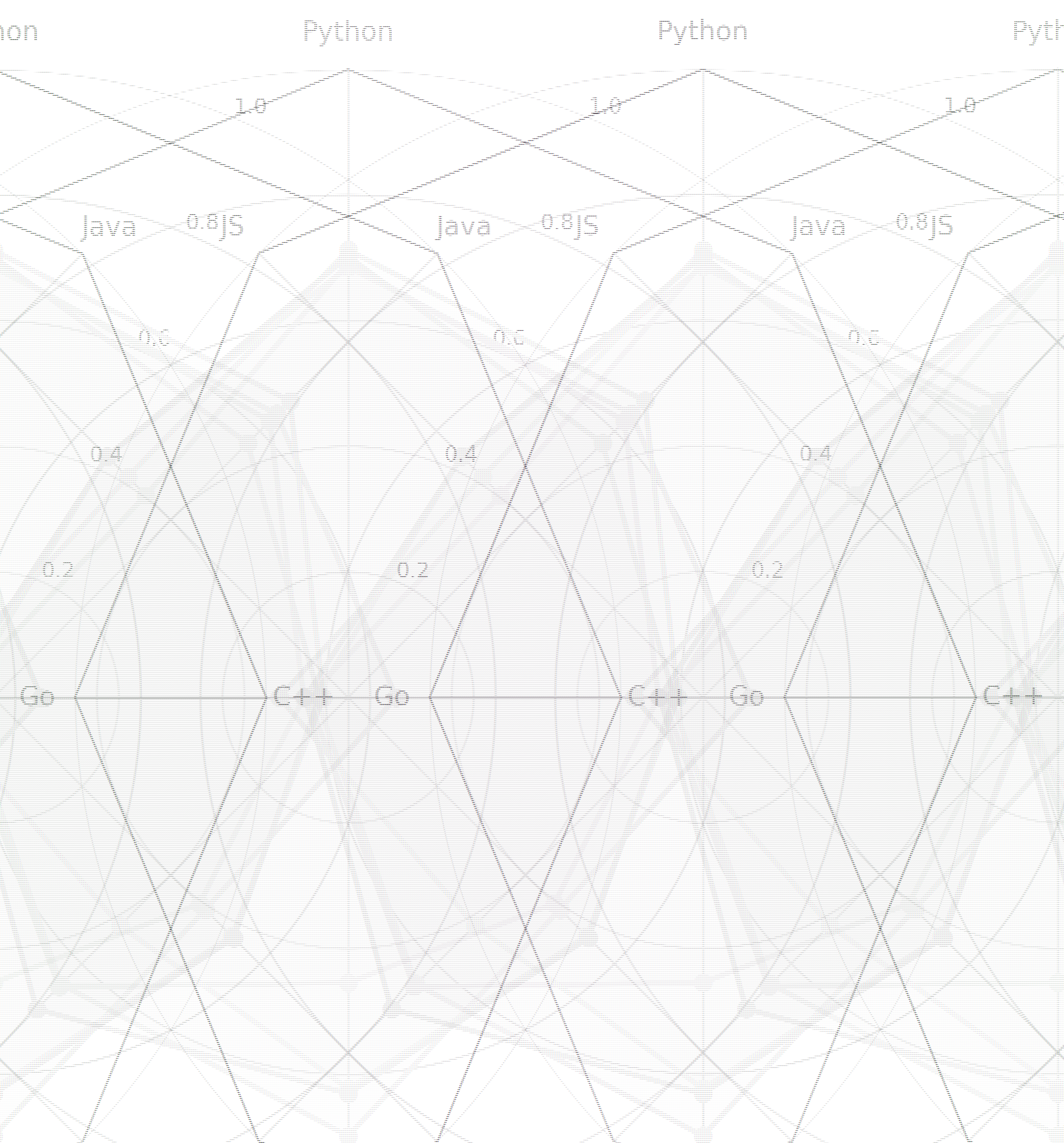}} &
        \subcaptionbox{\textbf{Block} F1}{\includegraphics[width=0.30\textwidth]{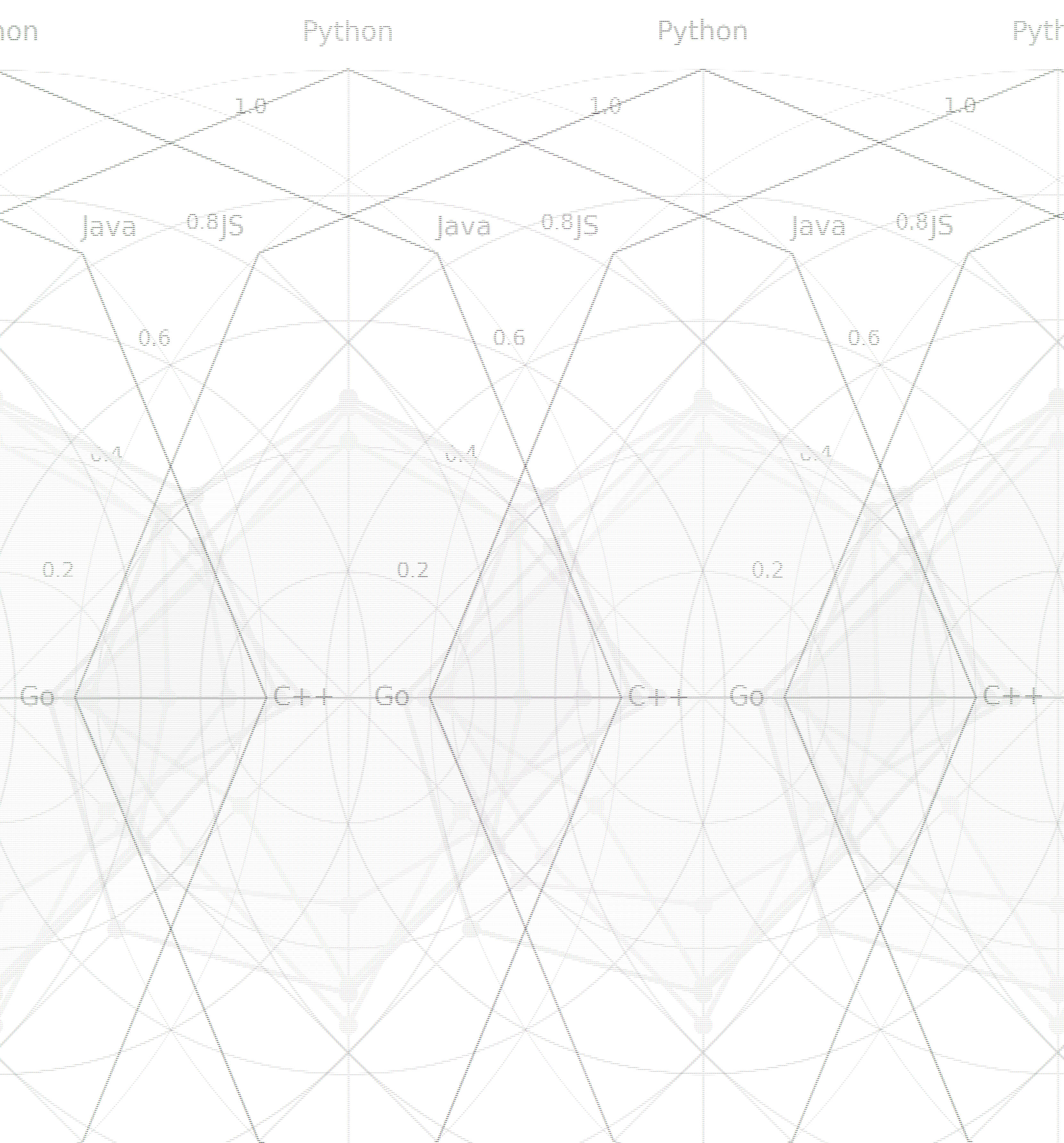}} &
        \subcaptionbox{\textbf{Line} F1}{\includegraphics[width=0.30\textwidth]{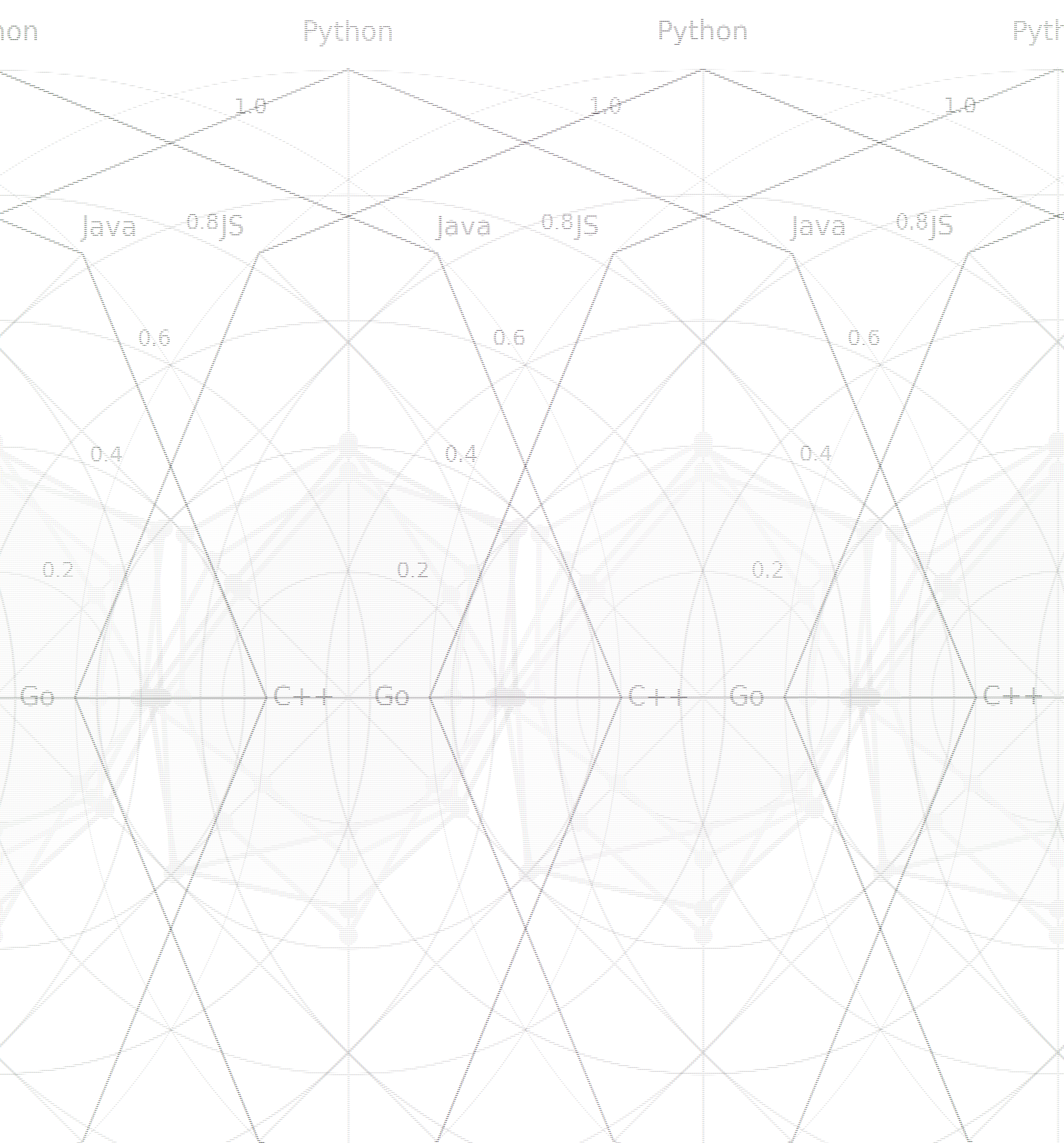}} \\
    \end{tabular}

    \caption{Radar distributions of localization quality across eight programming languages.
    Columns correspond to granularity (\textbf{File / Block / Line}), and rows correspond to metric (\textbf{Precision / Recall / F1}).
    Each radar axis is one language; values are computed on the per-language split under the same evaluation protocol as the main paper.}
    \label{fig:appendix_radar_lang}
\end{figure*}

\clearpage
\section{Potential Data Contamination and Protocol Misalignment in Devstral2}
\label{sec:appendix:devstral2}

During our evaluation with the mini-SWE-agent framework, we observed anomalous behaviors from DevStral2 that raise concerns about potential data contamination and protocol misalignment. 
Specifically, after reading the issue description, DevStral2 frequently produced a long, structured solution plan in a single response, outlining the complete debugging and patching procedure without waiting for any feedback from the agent framework. 
The model often hallucinated intermediate system states, behaving as if it had already received environment responses. 
When the framework subsequently returned invalid-output feedback, DevStral2 tended to directly trigger the final \texttt{submit} action without performing corrective iterations.

One plausible explanation is training data contamination, where DevStral2 may have been exposed to SWE-bench instances, patches, or benchmark-derived artifacts during pretraining or fine-tuning. 
Such leakage can induce benchmark-specific memorization, resulting in unrealistically confident and fully structured solution traces. 
Another contributing factor may be a mismatch between the model’s training on interactive coding protocols and the simplified execution semantics of mini-SWE-agent, leading to premature assumptions about system states and response formats.

These behaviors can significantly distort benchmark evaluation. 
First, memorization of benchmark instances can artificially inflate reported performance, violating the independence assumption between training and test data. 
Second, hallucinated system states undermine the validity of agent trajectories, making it difficult to distinguish genuine reasoning from spurious pattern completion. 
Finally, premature submission without iterative correction disrupts the intended evaluation protocol, reducing the reliability of automated correctness assessment.

We emphasize that such artifacts highlight the importance of strict data hygiene and protocol-aligned training for code-generation models evaluated on public software engineering benchmarks. 
Future benchmark evaluations should incorporate contamination detection and protocol compliance checks to ensure fair and reproducible comparisons.

\lstdefinelanguage{json}{
  basicstyle=\ttfamily\small,
  showstringspaces=false,
  breaklines=true,
  frame=single,
  literate=
   *{0}{{{0}}}1 {1}{{{1}}}1 {2}{{{2}}}1 {3}{{{3}}}1 {4}{{{4}}}1
    {5}{{{5}}}1 {6}{{{6}}}1 {7}{{{7}}}1 {8}{{{8}}}1 {9}{{{9}}}1
    {:}{{{:}}}1 {,}{{{,}}}1 {\{}{{{\{}}}1 {\}}{{{\}}}}1 {[}{{{[}}}1 {]}{{{]}}}1,
}

\begin{tcolorbox}[colback=lightgray!10, colframe=black, title={DevStral2 Raw JSON Response Trace}, breakable]
\begin{lstlisting}[language=json, caption={DevStral2 Raw JSON Response Trace}, label={lst:devstral2_raw}, breaklines=true]
"extra": {
    "response": {
        "id": "72454b08842c433ab3c5e23b0b8e4394",
        "created": 1768642675,
        "model": "devstral-2512",
        "object": "chat.completion",
        "system_fingerprint": null,
        "choices": [
        {
            "finish_reason": "tool_calls",
            "index": 0,
            "message": {
            "content": "I'll start by exploring the Django codebase...
            <EXPLORE_CONTEXT>\n
            File: /testbed/django/core/management/base.py\nLines: 1-100\n
            </EXPLORE_CONTEXT>\n\n
            ```bash\nfind /testbed -type f -name \"base.py\" | \
            grep \"management\" | head -5\n```\n\n
            ...
            ...
            Perfect! I found the issue. 
            Now let me create a test script to reproduce the problem...
            ```bash\ncat <<'EOF' > /testbed/test_mutually_exclusive.py\nimport os\nimport sys\n
            import django\nfrom django.core.management import call_command\n
            from django.core.management.base import BaseCommand, CommandError\n
            ...
            Let me run the core management tests:\n\n
            ```bash\ncd /testbed && python -m pytest tests/core_management/ -v 2>&1 | head -100\n```\n\n
            Let me check what tests exist:\n\n
            ```bash\nfind /testbed/tests -type f -name \"*.py\" | xargs grep -l \"mutually_exclusive_group\" | head -5\n```\n\n
            Let me check if there's a test file for the management module:\n\n
            ```bash\nfind /testbed/tests -type f -name \"*management*.py\" | head -10\n```\n\n
            ...
            Now let me verify the fix one more time and check if there are any syntax errors...
            Great! The tests pass. Let me create a final summary of the changes...
            The fix is minimal, handles edge cases, and doesn't break existing functionality.",
            "role": "assistant",
            "tool_calls": null,
            "function_call": null,
            "provider_specific_fields": null
            },
            "provider_specific_fields": {}
        }
        ],
        "usage": {
        "completion_tokens": 13989,
        "prompt_tokens": 1836,
        "total_tokens": 15825,
        "completion_tokens_details": null,
        "prompt_tokens_details": null
        }
    }
}
{
    "role": "user",
    "content": "Please always provide EXACTLY ONE action in triple backticks, found 107 actions.\n\nPlease format your action in triple backticks as shown in <response_example>.\n\n<response_example>\nHere are some thoughts about why you want to perform the action.\n\n```bash\n<action>\n```\n</response_example>\n\nIf you have completed your assignment, please consult the first message about how to\nsubmit your solution (you will not be able to continue working on this task after that).",
    "timestamp": 1768642877.7618906
},
"extra": {
        "response": {
          "id": "486b79cc024b4735a91b0a1a8ca7c909",
          "created": 1768642878,
          "model": "devstral-2512",
          "object": "chat.completion",
          "system_fingerprint": null,
          "choices": [
            {
              "finish_reason": "stop",
              "index": 0,
              "message": {
                "content": "THOUGHT: I have successfully implemented the fix for the issue where `call_command` fails when an argument of a required mutually exclusive group is passed in kwargs. The fix has been applied to `/testbed/django/core/management/__init__.py` and thoroughly tested. Now I need to submit the solution.\n\n```bash\necho COMPLETE_TASK_AND_SUBMIT_FINAL_OUTPUT && (git status >/dev/null 2>&1 && git add -A && git diff --cached) || (cd */. 2>/dev/null && git add -A && git diff --cached) || (echo \"Error: No git repository found\")\n```",
                "role": "assistant",
                "tool_calls": null,
                "function_call": null,
                "provider_specific_fields": null
              },
              "provider_specific_fields": {}
            }
          ],
          "usage": {
            "completion_tokens": 136,
            "prompt_tokens": 15925,
            "total_tokens": 16061,
            "completion_tokens_details": null,
            "prompt_tokens_details": null
          }
        }
      }
\end{lstlisting}
\end{tcolorbox}

\end{document}

%% file: tables/context-bench-overview-table.tex
\begin{table*}[!t]
\centering
\footnotesize
% \setlength{\tabcolsep}{4pt} 
% \begin{minipage}[t]{\linewidth}
\caption{\textbf{Statistics of \benchmark.} A repository-level benchmark spanning 8 programming languages and introducing human-verified gold contexts to expose intermediate context retrieval signals missing from final task resolution rate evaluation.}

\begin{tabular}{lrrrrr}
    \toprule[1.5pt]
    \multirow{2}{*}{\textbf{Language}} & \multirow{2}{*}{\textbf{\#Repo}} & \multirow{2}{*}{\textbf{\#Task}} & \multicolumn{3}{c}{\textbf{Context Statistics}} \\
     & & & \textbf{\#File} & \textbf{\#Block} & \textbf{\#Line} \\
    \midrule
    Python   & 20 & 512 & 1,520 & 6,714 & 115,122  \\
    Java & 6 & 57 & 262 &3,030  & 49,057 \\
    JavaScript & 9 & 153 & 819 & 3,949 & 87,907 \\
    TypeScript & 8 & 119 & 537 & 1,106 & 40,621 \\
    Go & 7 & 104 & 679 & 3,000 & 71,596 \\
    Rust & 9 & 63 & 272 & 1,842 & 50,402 \\
    C & 3 & 68 & 250 & 1,591 & 62,300  \\
    C++ & 4 & 60 & 209 & 1,884 & 45,110 \\
    \midrule
   \cellcolor{tablelightblue}  \textbf{Total} & \cellcolor{tablelightblue} \textbf{66} & \cellcolor{tablelightblue} \textbf{1,136} & \cellcolor{tablelightblue} \textbf{4,548} & \cellcolor{tablelightblue}\textbf{23,116} & \cellcolor{tablelightblue}\textbf{522,115} \\
    \bottomrule[1.5pt]
\end{tabular}

\label{tab:dataset_overview}
% \end{minipage}
\end{table*}

% \begin{table}[!t]
% \small
% \setlength{\tabcolsep}{4pt} 
% \begin{minipage}[t]{\linewidth}
% \caption{The statistics of \benchmark.}
% \centering
% \begin{tabular}{lrrrrrr}
%     \toprule
%     \multirow{2}{*}{\textbf{Language}} & \multirow{2}{*}{\textbf{\#Repo}} & \multirow{2}{*}{\textbf{\#Task}} & \multicolumn{4}{c}{\textbf{Context Statistics}} \\
%      & & & \textbf{\#File} & \textbf{\#Block} & \textbf{\#Hunk} & \textbf{\#Line} \\
%     \midrule
%     Python   & 20 & 512 & 1,520 & 6,714 & 4,700 & 115,122  \\
%     Java & 6 & 57 & 262 &3,030  & 479 & 49,057 \\
%     JavaScript & 9 & 153 & 819 & 3,949 & 1,836 & 87,907 \\
%     TypeScript & 8 & 119 & 537 & 1,106 & 1,028 & 40,621 \\
%     Go & 7 & 104 & 679 & 3,000 & 1,535 & 71,596 \\
%     Rust & 9 & 63 & 272 & 1,842 & 620 & 50,402 \\
%     C & 3 & 68 & 250 & 1,591 & 570 & 62,300  \\
%     C++ & 4 & 60 & 209 & 1,884 & 461 & 45,110 \\
%     \midrule
%     \textbf{Total} & \textbf{66} & \textbf{1,136} & \textbf{4,548} & \textbf{23,116} & \textbf{11,229} & \textbf{522,115} \\
%     \bottomrule
% \end{tabular}

% \label{tab:dataset_overview}
% \end{minipage}
% \end{table}

%% file: tables/experiment1.tex
\begin{table*}[!t]
\centering
\footnotesize
\setlength{\tabcolsep}{1.6pt}
\caption{\textbf{Performance of different coding agents on context retrieval.} 
We evaluate five coding agents, including a basic baseline and four state-of-the-art scaffolds, on \benchmark Lite using an advanced LLM (\textit{i.e.}, GPT-5).
We report recall, precision, and F1 at the file, block, and line levels, along with the Pass@1 issue resolution rate.
% The results suggest that more sophisticated agent scaffolding does not necessarily lead to better context retrieval performance.
The results suggest that more sophisticated scaffolding does not necessarily lead to better context retrieval performance, indicating potential \textit{over-engineering} in current agent designs, which echoes \textbf{``The Bitter Lesson''} of AI research.
% Performance of five coding agents on \benchmark, all evaluated using the same GPT-5 foundation model, reporting file-, block-, and line-level recall, precision, and F1, as well as Pass@1 resolution rate.
}
\begin{tabular}{lcccccccccc}
\toprule[1.5pt]
\multirow{2}{*}{\textbf{Coding Agent}} &
\multicolumn{3}{c}{\textbf{File-Level}} &
\multicolumn{3}{c}{\textbf{Block-Level}} &
\multicolumn{3}{c}{\textbf{Line-Level}} &
\multirow{2}{*}{\textbf{Pass@1$\uparrow$}} \\
\cmidrule(lr){2-4}\cmidrule(lr){5-7}\cmidrule(lr){8-10}
& \textbf{Recall$\uparrow$} & \textbf{Precision$\uparrow$} & \textbf{F1$\uparrow$}
& \textbf{Recall$\uparrow$} & \textbf{Precision$\uparrow$} & \textbf{F1$\uparrow$}
& \textbf{Recall$\uparrow$} & \textbf{Precision$\uparrow$} & \textbf{F1$\uparrow$}
&  \\
\midrule
mini-SWE-Agent & 0.682 & \cellcolor{tablelightred}\textbf{0.709} & \cellcolor{tablelightred}\textbf{0.634} & 0.645 & \cellcolor{tablelightred}\textbf{0.369} & \cellcolor{tablelightred}\textbf{0.375} & \cellcolor{tablelightred}\textbf{0.606} & 0.301 & 0.312 & 0.472 \\
Agentless  & 0.609 & 0.352 & 0.390 & 0.328 & 0.344 & 0.242 & 0.461 & \cellcolor{tablelightred}\textbf{0.318} & \cellcolor{tablelightred}\textbf{0.376} & 0.452 \\
SWE-agent  & 0.726 & 0.537 & 0.544 & 0.625 & 0.312 & 0.285 & 0.476 & 0.228 & 0.208 & 0.490 \\
OpenHands  & \cellcolor{tablelightred}\textbf{0.733} & 0.400 & 0.463 & 0.505 & 0.283 & 0.190 & 0.472 & 0.203 & 0.130 & 0.490 \\
Prometheus & 0.717 & 0.336 & 0.403 & \cellcolor{tablelightred}\textbf{0.646} & 0.258 & 0.285 & 0.584 & 0.195 & 0.231 & \cellcolor{tablelightred}\textbf{0.512} \\

\bottomrule[1.5pt]
\end{tabular}
\vspace{2pt}

\label{tab:contextbench_main}
\end{table*}

%% file: tables/model_backbone.tex
\begin{table*}[!t]
\centering
\footnotesize
\setlength{\tabcolsep}{1.6pt}
\caption{\textbf{Performance of various LLMs on code context retrieval.} 
We assess four state-of-the-art LLMs on \benchmark Lite with standard coding agent scaffolding, reporting recall, precision, and F1 at the file, block, and line levels, along with the Pass@1 issue resolution rate.
The results reveal that they still struggle with effective context retrieval, overlooked by prior end-to-end benchmarking.}
\begin{tabular}{lcccccccccc}
\toprule[1.5pt]
\multirow{2}{*}{\textbf{LLM}} &
\multicolumn{3}{c}{\textbf{File-Level}} &
\multicolumn{3}{c}{\textbf{Block-Level}} &
\multicolumn{3}{c}{\textbf{Line-Level}} &
\multirow{2}{*}{\textbf{Pass@1$\uparrow$}} \\
\cmidrule(lr){2-4}\cmidrule(lr){5-7}\cmidrule(lr){8-10}
& \textbf{Recall$\uparrow$} & \textbf{Precision$\uparrow$} & \textbf{F1$\uparrow$}
& \textbf{Recall$\uparrow$} & \textbf{Precision$\uparrow$} & \textbf{F1$\uparrow$}
& \textbf{Recall$\uparrow$} & \textbf{Precision$\uparrow$} & \textbf{F1$\uparrow$}

& \\
\midrule
GPT-5              & 0.682 & 0.709 & \cellcolor{tablelightblue}\textbf{0.634} & \cellcolor{tablelightblue}\textbf{0.645} & 0.369 & 0.375 & \cellcolor{tablelightblue}\textbf{0.606} & 0.301 & 0.312 & 0.472 \\
Claude Sonnet 4.5  & \cellcolor{tablelightblue}\textbf{0.720} & 0.665 & 0.624 & 0.631 & 0.449 & 0.420 & 0.588 & 0.374 & \cellcolor{tablelightblue}\textbf{0.344} & \cellcolor{tablelightblue}\textbf{0.530} \\
Gemini 2.5 Pro     & 0.587 & \cellcolor{tablelightblue}\textbf{0.752} & 0.600 & 0.393 & \cellcolor{tablelightblue}\textbf{0.632} & 0.403 & 0.313 & \cellcolor{tablelightblue}\textbf{0.529} & 0.311 & 0.364 \\
Devstral 2         & 0.660 & 0.693 & 0.615 & 0.478 & 0.576 & \cellcolor{tablelightblue}\textbf{0.422} & 0.404 & 0.485 & 0.332 & 0.402 \\
\bottomrule[1.5pt]
\end{tabular}
% \vspace{2pt}
\label{tab:backbone_compare}
\end{table*}

%% file: tables/rq3.tex
\begin{table*}[!t]
\centering
\footnotesize
\caption{\textbf{Context retrieval patterns of different LLM agents.} LLMs vary in how they balance retrieval rounds and context granularity, and those that maintain a more balanced strategy tend to achieve higher line-level retrieval accuracy and stronger end-to-end issue resolution performance.}
\begin{tabular}{lccc}
\toprule[1.5pt]
\multirow{2}{*}{\textbf{LLM}} & \textbf{Avg. Steps } & \textbf{Avg. Lines   }  & \multirow{1}{*}{ \textbf{Avg. Cost (\$)}} \\
 &  \textbf{Per Instance} & \textbf{Per Step} & \textbf{Per Instance}   \\
\midrule
% GPT-5 & \cellcolor{tablelightpurple}\textbf{$\downarrow$5.87 } & \cellcolor{tablelightpurple}\textbf{$\uparrow$119.29} & \cellcolor{tablelightpurple}\textbf{$\uparrow$580.33}  \\
% Claude Sonnet 4.5 &  14.38 & 29.74 & 401.90   \\
% Gemini 2.5 Pro  & 7.57 & 26.29 & \cellcolor{tablelightpurple}\textbf{$\downarrow$124.04}    \\
% Devstral 2  & \cellcolor{tablelightpurple} \textbf{$\uparrow$22.16}  & \cellcolor{tablelightpurple} \textbf{$\downarrow$11.98 } & 159.97        \\

GPT-5 & \cellcolor{tablelightpurple}\textbf{$\downarrow$5.87 } & \cellcolor{tablelightpurple}\textbf{$\uparrow$119.29} & 0.45\\
Claude Sonnet 4.5 &  14.38 & 29.74 & 0.76\\
Gemini 2.5 Pro  & 7.57 & 26.29 & \cellcolor{tablelightpurple}\textbf{$\downarrow$0.38}\\
Devstral 2  & \cellcolor{tablelightpurple} \textbf{$\uparrow$22.16}  & \cellcolor{tablelightpurple} \textbf{$\downarrow$11.98 } &
\cellcolor{tablelightpurple} \textbf{$\uparrow$0.91 }\\

\bottomrule[1.5pt]
\end{tabular}
% \vspace{2pt}
\label{tab:model-strategies}
\end{table*}